\documentclass{article}
\usepackage[utf8]{inputenc}
\usepackage[T1]{fontenc}
\usepackage{geometry}
\usepackage{translator}
\usepackage[catalan, spanish, british]{babel}
\usepackage{johd}

\usepackage{titling}
\pretitle{}
\posttitle{}
\preauthor{}
\postauthor{}
\predate{}
\postdate{}

\renewcommand{\maketitlehooka}{%
  \begin{center}%
}
\renewcommand{\maketitlehookb}{%
  \end{center}%
  \vspace{1em}%
}

\newcommand{\subtitle}[1]{%
  \gdef\@subtitle{#1}%
}
\newcommand{\printsubtitle}{%
  {\large\emph{\@subtitle}\par}%
  \vspace{0.5em}%
}

\usepackage[colorinlistoftodos]{todonotes}

\usepackage{hyperref}
\usepackage{soul}
\usepackage{underscore}
\usepackage{array, multirow}
\usepackage{tabularx, tabulary, longtable}
\usepackage{tipa}
\usepackage{multicol}
\usepackage{makecell}
\usepackage{booktabs}
\usepackage{enumitem}
\usepackage{ragged2e}
\usepackage{xspace}
\usepackage{setspace}
\usepackage{etoolbox}

\usepackage{pgfplots}
\usepackage{pgfplotstable}
\pgfplotsset{compat=1.18}

\newcolumntype{Y}{>{\RaggedRight\arraybackslash}X}

\let\oldbibliography\thebibliography
\renewcommand{\thebibliography}[1]{%
  \oldbibliography{#1}%
  \setlength{\itemsep}{8pt}%
  \setlength{\baselineskip}{11.5pt}%
}

\newcommand{\mapu}[1]{\emph{#1}}
\newcommand{\mapuex}[1]{\footnotesize\emph{#1}}
\newcommand{\gloss}[1]{\footnotesize\textsc{\lowercase{#1}}}
\newcommand{\example}[1]{\footnotesize\textbf{E.#1}\xspace}

\usepackage{titlesec}
\titleformat{\section}
  {\normalfont\Large\bfseries}
  {\thesection}{1em}{}
\titleformat{\subsection}
  {\normalfont\large\bfseries}
  {\thesubsection}{1em}{}
\titleformat{\subsubsection}
  {\normalfont\normalsize\bfseries}
  {\thesubsubsection}{1em}{}

\hypersetup{
    colorlinks=true,
    linkcolor=blue,
    citecolor=violet,
    urlcolor=blue,
    pdftitle={Preverbal Uninflected and Underived Roots in \mapu{Mapudüngun}},
    pdfauthor={Andrés Chandía}
}

\title{\textbf{Preverbal Uninflected and Underived Roots in \mapu{Mapudüngun}}}
\subtitle{\Large \emph{Wüño} and Its Implications}
\author{%
Andrés Chandía\\[2pt]
\small\href{mailto:andres@chandia.net}{\texttt{andres@chandia.net}}\\[8pt]
Department of Catalan Philology and General Linguistics\\ 
University of Barcelona\\[4pt]
PhD programme: Cognitive Science and Language\\[12pt]
{\small Supervised by}\\
{\small Dr. Elisabet Comelles \href{mailto:elicomelles@ub.edu}{\texttt{(elicomelles@ub.edu)}} and\\
{\small Dr. Irene Castellón \href{mailto:icastellon@ub.edu}{\texttt{(icastellon@ub.edu)}}}}
}
\date{}

\begin{document}

% ========== TITLE PAGE ==========
% Custom title page to avoid titling package issues
\begin{center}
  \vspace*{2cm}
  {\Huge\bfseries Preverbal Uninflected and Underived Roots in \mapu{Mapudüngun}\par}
  \vspace{1em}
  {\Large\emph{Wüño} and Its Implications\par}
  \vspace{2cm}
  {\large
  Andrés Chandía\\[2pt]
  \small\href{mailto:andres@chandia.net}{\texttt{andres@chandia.net}}\\[8pt]
  Department of Catalan Philology and General Linguistics\\ 
  University of Barcelona\\[4pt]
  PhD programme: Cognitive Science and Language\\[12pt]
  {\small Supervised by}\\
  {\small Dr. Elisabet Comelles \href{mailto:elicomelles@ub.edu}{\texttt{(elicomelles@ub.edu)}} and\\
  {\small Dr. Irene Castellón \href{mailto:icastellon@ub.edu}{\texttt{(icastellon@ub.edu)}}}}
  }
  \vfill
\end{center}
\thispagestyle{empty}

\newpage
\begin{abstract}
\noindent This study examines the grammatical status of preverbal uninflected and underived roots in \mapu{Mapudüngun}, with particular focus on \mapu{wüño} `return/re-'. Through a critical review of scholarly classifications—auxiliaries \citep{smeets2008}, modal prefixes \citep{longkon2011}, and preverbal particles/complex verb stems \citep{zuniga2006}—we demonstrate the limitations of existing frameworks. A diachronic corpus analysis spanning four centuries (1606–present) reveals that these elements exhibit three distinct profiles: stable V1 compounds (\mapu{kim, shinge}), volatile V1 rates reflecting orthographic shift (\mapu{pepi, wüño}), and a true particle (\mapu{kalli}). The discovery of V2 attestations for \mapu{kim} and \mapu{küpa} confirms their status as full lexical verbs. We propose a prosodic-orthographic hypothesis: apparent ``variable binding'' results from the fossilization of prosodic pauses transcribed by early missionaries as spaces, a convention later reanalysed by speakers as syntactic boundaries. The evidence supports \cite{zuniga2006}'s radical concatenation as the correct grammatical model, with implications for the study of languages with no pre-contact written tradition.
\end{abstract}


\selectlanguage{spanish}
\begin{abstract}
\noindent Este estudio examina el estatus gramatical de las raíces preverbales no flexionadas y no derivadas en \mapu{mapudüngun}, con especial atención a \mapu{wüño} `volver/re-'. Mediante una revisión crítica de las clasificaciones académicas—auxiliares \citep{smeets2008}, prefijos modales \citep{longkon2011} y partículas preverbales/verbos complejos \citep{zuniga2006}—demostramos las limitaciones de los marcos existentes. Un análisis de corpus diacrónico que abarca cuatro siglos (1606–presente) revela que estos elementos presentan tres perfiles diferenciados: compuestos V1 estables (\mapu{kim, shinge}), tasas V1 volátiles que reflejan cambios ortográficos (\mapu{pepi, wüño}) y una partícula genuina (\mapu{kalli}). El descubrimiento de atestaciones en posición V2 para \mapu{kim} y \mapu{küpa} confirma su estatus como verbos léxicos completos. Proponemos una hipótesis prosódico-ortográfica: la aparente ``vinculación variable'' resulta de la fosilización de pausas prosódicas transcritas por los primeros misioneros como espacios, una convención posteriormente reanalizada por los hablantes como límites sintácticos. La evidencia respalda la concatenación radical de \cite{zuniga2006} como modelo gramatical correcto, con implicaciones para el estudio de lenguas sin tradición escrita previa al contacto.
\end{abstract}

\noindent\textbf{Palabras clave:} Mapuche; Mapudüngun; mofosintaxis; afijos; verbos; prefijos modales; elementos preverbales; hipótesis prosódico-ortográfica; concatenación radical; contacto lingüístico; lingüística de corpus diacrónica

\selectlanguage{catalan}
\begin{abstract}
\noindent Aquest estudi examina l'estatus gramatical de les arrels preverbals no flexionades i no derivades en \mapu{mapudüngun}, amb especial atenció a \mapu{wüño} `tornar/re-'. Mitjançant una revisió crítica de les classificacions acadèmiques—auxiliars \citep{smeets2008}, prefixos modals \citep{longkon2011} i partícules preverbals/verbs complexos \citep{zuniga2006}—demostrem les limitacions dels marcs existents. Una anàlisi de corpus diacrònic que abasta quatre segles (1606–present) revela que aquests elements presenten tres perfils diferenciats: compostos V1 estables (\mapu{kim, shinge}), taxes V1 volàtils que reflecteixen canvis ortogràfics (\mapu{pepi, wüño}) i una partícula genuïna (\mapu{kalli}). El descobriment d'atestacions en posició V2 per a \mapu{kim} i \mapu{küpa} confirma el seu estatus com a verbs lexicals complets. Proposem una hipòtesi prosòdico-ortogràfica: l'aparent ``vinculació variable'' resulta de la fossilització de pauses prosòdiques transcrites pels primers missioners com a espais, una convenció posteriorment reanalitzada pels parlants com a límits sintàctics. L'evidència recolza la concatenació radical de \cite{zuniga2006} com a model gramatical correcte, amb implicacions per a l'estudi de llengües sense tradició escrita pre-contacte.
\end{abstract}

\noindent\textbf{Paraules clau:} Maputxe; Mapudüngun; mofosintaxis; afixos; verbs; prefixos modals; elements preverbals; hipòtesi prosòdico-ortogràfica; concatenació radical; contacte lingüístic; lingüística de corpus diacrònica

\newpage
% ========== MAIN CONTENT ==========
\selectlanguage{british}

% ========== TABLE OF CONTENTS ==========
\tableofcontents

\newpage
% ========== NOTE ON MAPUCHE NAMES (REVINDICATIVE) ==========
\begin{center} \begin{minipage}{14cm}
\footnotesize \textbf{Note on the spelling of \mapu{Mapuche} names}: In this article, we spell \mapu{Mapuche} names and surnames according to the \textit{Alfabeto Mapuche Unificado} (AMU), as an act of recognition and respect towards the \mapu{Mapuche} people and their language. For example, we use the forms \mapu{Longkon} and \mapu{Koña} instead of the standard bibliographical Castilianisations (Loncón, Coña). In the bibliographic references, however, the standardised forms are added in parentheses to facilitate the location of the sources.
\normalsize
\end{minipage}\end{center}

\vspace{1cm}
\section{\label{sec.01} Introduction}

The verbal morphology of \mapu{Mapudüngun} (\mapu{Mapuche}, isolate) has long attracted the attention of linguists due to its polysynthetic character and complex system of valence-adjusting operations (\citealp{smeets2008}; \citealp{zuniga2006}). Among the most debated aspects of the language's verbal compound is a set of uninflected and underived roots that appear in preverbal position, modifying the meaning of the following verb root. These elements—including \mapu{kalli} `let, allow', \mapu{kim} `know how', \mapu{küpa} `want', \mapu{pepi} `can', \mapu{shinge} `move along', and \mapu{wüño} `return/re-'—have received divergent classifications in the scholarly literature, reflecting broader theoretical disagreements about the boundary between morphology and syntax, and between lexical and grammatical categories.

The central question addressed by this study is simple to state but surprisingly difficult to answer: what kind of linguistic objects are these preverbal roots? Are they auxiliaries, as proposed by \cite{smeets2008}? Modal prefixes, as argued by \cite{longkon2011}? Preverbal particles or members of complex verb stems, as suggested by \cite{zuniga2006}? Or does their hybrid behaviour resist classification within any of these established categories?

The stakes of this question extend beyond the accurate description of \mapu{Mapudüngun}. At issue are fundamental assumptions about how we identify grammatical categories in languages with no pre-contact written tradition, how we interpret orthographic variation in historical sources, and how we distinguish between grammatical change and the mere appearance of change created by shifting conventions of writing.

This study addresses these questions through a combination of critical literature review and original corpus analysis. Section \ref{sec.02} proposes a unified prosodic-orthographic hypothesis that reframes the apparent contradictions. Section \ref{sec.03} describes the diachronic corpus used to test this hypothesis. Section \ref{sec.04} presents the quantitative results, and Section \ref{sec.05} discusses their implications for our understanding of \mapu{Mapudüngun} grammar and for the study of language contact and orthographic change more broadly.

A note on translation is necessary before proceeding. The glosses assigned to these elements in the literature – particularly \mapu{küpa} as `want' – often reflect Eurocentric interpretations rather than indigenous semantic categories. \mapu{Mapudüngun} speakers frequently use the motion verb \mapu{küpa} `to come' in constructions that Spanish translators render as expressions of volition (e.g., \mapu{küpa langümfiñ} lit. `I come to kill him' → `I want to kill him'). This pattern, documented in \cite{guevara1913}, suggests that the `volitional' reading may be an artifact of translation rather than a grammaticalised modal category. Throughout this review, we retain conventional glosses for comparability with previous scholarship, but return to this issue in \S\ref{sec.04.5}, where the insertion evidence provides independent confirmation that these elements retain their core lexical semantics.

\subsection{\label{sec.01.1} Classifying Preverbal Uninflected and Underived Roots in \mapu{Mapudüngun}: A Review of Scholarly Approaches to Their Role in Verb Formation}

This section reviews scholarly approaches to preverbal elements in \mapu{Mapudüngun} – including \mapu{kalli} `let, allow', \mapu{kim} `know how', \mapu{küpa} `want', \mapu{pepi} `can', \mapu{shinge} `move along', and \mapu{wüño} `return, re-' – and their role in verb formation. It evaluates competing analyses that classify them as auxiliaries (\S\ref{sec.01.1.1}), modal prefixes (\S\ref{sec.01.1.2}), and preverbal particles or complex verb stems (\S\ref{sec.01.1.3}).

Regarding the \mapu{Mapudüngun} elements under analysis, \cite{smeets2008} classifies them as auxiliaries (\S\ref{sec.01.1.1}). This category is defined by \cite{moreno2000}, \cite{bybee1994}, and \cite{payne1997} as grammatical markers that host inflection and govern verb phrases (e.g., English \textit{must}). A crucial cross-linguistic reference for this discussion is \cite{anderson2006}, which provides a comprehensive typological overview of auxiliary verb constructions, establishing the range of variation and the diagnostic criteria commonly used to identify them.

However, \cite{smeets2008}'s identification of five `auxiliaries' is complicated by their dual use as independent verbs. Key contradictions include:
\begin{itemize}
\item failure to meet inflection-hosting criteria (e.g., negation/imperative attaches to the main verb: \mapu{pepi küdaw-\textbf{la}-n} `I cannot work');
\item retention of lexical meaning (\mapu{küpa} `want/come' contrasts with grammaticalised auxiliaries).
\end{itemize}

These issues suggest that the auxiliary label is problematic due to the elements' hybrid behaviour (particularly their lexical retention and variable binding).

The alternative classification of ``modal prefixes'' by \cite{longkon2011} (\S\ref{sec.01.1.2}) is evaluated against definitions from \cite{dixon2010} and \cite{aikhenvald2006}, who describe them as bound morphemes expressing modality (e.g., Tariana\footnote{Tariana is an Arawak language spoken in the Vaupés region of northwestern Brazil. The modal suffix \textit{-kade-} `can' is described by \cite{aikhenvald2006} as a bound morpheme that attaches directly to the verb root and cannot occur independently. This stands in contrast to the \mapu{Mapudüngun} elements under discussion, which—as we demonstrate—retain independent verbal properties and can host inflection. The Tariana example is invoked here to illustrate the typological expectations of modal prefixes, not to suggest a direct parallel with \mapu{Mapudüngun}.} \textit{-kade-} `can').

While \cite{longkon2011} analyses these elements as prefixes, she notes the separability of \mapu{kalli}. This analysis is challenged by:
\begin{itemize}
\item Variable binding (\mapu{kalli düngu-pe!} `let him speak!' vs. fused \mapu{pepi-weyel-} `can swim')
\item Orthographic variation, which reflects speaker intuition (pointing to clitic-like gradience)
\end{itemize}

Thus, the prefix label appears oversimplified; the evidence suggests these elements behave more like clitics or V1 in compounds.

The review then turns to \cite{zuniga2006}'s framework, which categorises the elements as either ``preverbal particles'' or ``complex verb stems (CVS)'' (\S\ref{sec.01.1.3}). These terms are defined by \cite{payne1997} and \cite{baker1996} as follows:
\begin{itemize}
\item Preverbal particles: Free morphemes in fixed preverbal position (e.g., English \textit{must})
\item CVS: Compounds of verb roots (e.g., Yimas\footnote{Yimas is a Lower Sepik-Ramu language spoken in Papua New Guinea. The compound \textit{kpa-ŋka-} `want-eat' \citep{baker1996} illustrates the typological pattern of verb-verb compounding where the first root (V1) modifies the second (V2). This is directly parallel to the structure we argue for \mapu{Mapudüngun}: the preverbal elements are V1 modifiers in a productive compounding system, not a special grammatical class.\\} \textit{kpa-ŋka-} `want-eat')
\end{itemize}

\cite{zuniga2006}'s analysis treats \mapu{kim} and \mapu{pepi} as both particles (\mapu{kim weyel-}) and CVS (\mapu{kim-weyel-}). However, this approach is problematic because:
\begin{itemize}
\item CVS assumes unitary behaviour, yet particles retain independent meanings (\mapu{kim} `wise')
\item Orthographic inconsistency (\mapu{pepi weyel-} vs. \mapu{pepi-weyel-}) indicates clitic-like behaviour
\end{itemize}

Consequently, a hybrid analysis (classifying them as clitic/modifiers) better accounts for the continuum between fusion and separation.

Finally, earlier conservative classifications are examined, including bare ``prefixes'' (\citealp{deaugusta1903}; and de Augusta's dictionary in \citealp{chandia2014}) and bare ``modals'' \citep{salas1992a} (\S\ref{sec.01.1.4}). This section incorporates \mapu{wüño} into the analysis, drawing on de Augusta's definitions.

The behaviour of \mapu{wüño} (`return/re-') exhibits duality:
\begin{itemize}
\item Prefix-like when fused (\mapu{wüño-kintu-n} `look back')
\item Particle-like when separated (\mapu{wüño tu-n} `recover')
\end{itemize}

However, \mapu{wüño} fails to meet prefix criteria because it can host suffixes (\mapu{wüño-me-n} `I returned [there]'). Moreover, native orthography often favors separation, suggesting perceptual autonomy. Therefore, \mapu{wüño} is best analysed as a preverbal clitic/modifier, reconciling de Augusta's prefix label with contemporary data and aligning with typological parallels like Tariana's \textit{-kade} \citep{aikhenvald2006}.

\subsubsection{\label{sec.01.1.1} Auxiliaries  \citep{smeets2008}}

The concept of auxiliary verbs is recognised across different languages, although its precise definition varies according to structural behaviour. Three seminal perspectives have established key diagnostic criteria.

Following \cite{moreno2000}, an auxiliary is a grammatical marker that carries all inflection (tense, person, mood) while the main verb becomes invariant. The auxiliary governs the syntactic construction, as illustrated by English \textit{do}-support (\emph{She does not sing-Ø}).

\cite{bybee1994} propose that auxiliaries are grammaticalised verbs forming a closed class that expresses tense-aspect-modality (TAM), exhibiting asymmetric dependency with the main verb (e.g., Spanish \textit{\textbf{haber}} in \textit{\textbf{he} cantado} `I have sung'). Auxiliaries emerge from lexical verbs through grammaticalisation, undergoing phonetic reduction and syntactic dependency.

\cite{payne1997} further posits that prototypical auxiliaries exhibit phonological binding to the main verb, absorbing its inflection and occupying a fixed position within the sentence (e.g., Swahili\footnote{Swahili is a Bantu language spoken in East Africa. The prefix \textit{-na-} marks present tense and attaches directly to the verb root, hosting inflection and forming a single phonological word \citep{payne1997}. This exemplifies the prototypical auxiliary behaviour—obligatory boundness and inflectional absorption—that the \mapu{Mapudüngun} elements do not exhibit.} \textit{\textbf{-na-}} in \textit{a\textbf{-na-}soma} `s/he is reading').

According to \citeauthor{smeets2008} (\citeyear{smeets2008}: 175), \mapu{Mapudüngun} features five `auxiliary verbs', characterised by their position, non-derived form, and function. These verbal roots lack derivation and/or inflection, and directly precede the main verb without intervening elements or morphological/phonological binding to the main verbal form. However, these roots can also serve as the foundation for inflected verbal forms, meaning they can function as regular \mapu{Mapuche} verbs. Examples (\hyperref[e01]{E.01}–\hyperref[e05]{E.05}) illustrate this duality; in each pair, (a) presents the root as an `auxiliary', (b) as an inflected verbal root.

\paragraph{\example{01} \label{e01} \mapuex{kalli} `enabling'}~
\vspace{-10pt}\begin{enumerate}[label=\alph*.]
\item \label{e01a} \mapuex{kalli \hspace{2pt} düngu  \hspace{19pt}  -pe} \hfill (\citealp{smeets2008}: 176; 95)\\
      \gloss{AUX NN.matter +IMP.3}\\
      `let him speak!'
\item \label{e01b} \mapuex{kalli \hspace{56pt} -ka \hspace{7pt} -w \hspace{8pt} -küdaw \hspace{7pt} -küle -y \hspace{10pt} -ø \hspace{2pt} -iñ} \hfill (\citealp{smeets2008}: 452; 17)\\
      \gloss{IV.be-by-oneself +FAC +REF NN.work +ST +IND +1 +PL}\\
      `we let each other do our job'
\end{enumerate}
\normalsize

A curious aspect of (\hyperref[e01a]{E.01.a}) is that the imperative morpheme attaches not to the `auxiliary' \mapu{kalli} `let', but to the noun-derived action \mapu{düngu} `speak'. This appears to contradict \cite{smeets2008}'s proposal, as the translation suggests the exhortation is directed toward the second person, while the imperative affix \mapu{-pe} relates to the third person executing the action. By this reasoning, \mapu{kalli} should be bound, forming a verb stem with \mapu{düngu}.

\paragraph{\example{02} \label{e02} \mapuex{kim} `knowing how to'}~
\vspace{-10pt}\begin{enumerate}[label=\alph*.]
\item \label{e02a} \mapuex{kim \hspace{1pt} tuku \hspace{12pt} -fi \hspace{8pt} -n} \hfill (\citealp{smeets2008}: 175; 93)\\
      \gloss{AUX TV.put +3P +IND.1SG}\\
      `I know how to put it'
\item \label{e02b} \mapuex{kim \hspace{27pt} -tuku \hspace{10pt} -fi \hspace{8pt} -n} \hfill (\citealp{smeets2008}: 176; 96)\\
      \gloss{AJ.knower TV.put +3P +IND.1SG}\\
      `I gradually got to understand/know it'
\end{enumerate}
\normalsize

\paragraph{\example{03} \label{e03} \mapuex{küpa} `wishing'}~
\vspace{-10pt}\begin{enumerate}[label=\alph*.]
\item \label{e03a} \mapuex{küpa \hspace{10pt} pu \hspace{46pt} -le \hspace{6pt} -n \hspace{26pt} liwen} \hfill (\citealp{smeets2008}: 175; 92)\\
      \gloss{AUX TV.arrive-there +ST +IND.1SG NN.morning}\\
      `I want to arrive in the morning'
\item \label{e03b} \mapuex{küpa \hspace{18pt} -nu \hspace{6pt} -l \hspace{12pt} -i \hspace{4pt} -iñ} \hfill (\citealp{smeets2008}: 43; n)\\
      \gloss{IV.come +NEG +SBJ +1 +PL}\\
      `if we do not come'
\end{enumerate}
\normalsize

\paragraph{\example{04} \label{e04} \mapuex{pepi} `being able / can'}~
\vspace{-10pt}\begin{enumerate}[label=\alph*.]
\item \label{e04a} \mapuex{pepi \hspace{2pt} küdaw \hspace{10pt} -la \hspace{10pt} -n} \hfill (\citealp{smeets2008}: 175; 91)\\
      \gloss{AUX NN.work +NEG +IND.1SG}\\
      `I am not able to work'
\item \label{e04b} \mapuex{pepi \hspace{12pt} -l \hspace{10pt} -nge \hspace{10pt} -nu \hspace{6pt} -el} \hfill (\citealp{smeets2008}: 201; 104)\\
      \gloss{IV.can +CA +IV.be +NEG +OVN}\\
      `what cannot be handled'
\end{enumerate}
\normalsize

In (\hyperref[e04a]{E.04.a}), negation attaches to the verbalised root `work', not to the \textbf{auxiliary} `can'. Consequently, negation expresses absence of the action `work', rather than inability to work. An alternative formulation would be `I can [choose] not to work'. This approach parallels (\hyperref[e01a]{E.01.a}), where compounding appears more natural.

\paragraph{\example{05} \label{e05} \mapuex{shinge} `moving up/along'}~
\vspace{-10pt}\begin{enumerate}[label=\alph*.]
\item \label{e05a} \mapuex{pichi \hspace{18pt} shinge \hspace{2pt} amu \hspace{4pt} -m \hspace{6pt} -ø \hspace{3pt} -ün} \hfill (\citealp{smeets2008}: 175; 94)\\
      \gloss{AJ.little AUX \hspace{5pt} IV.go +IMP +2 +PL}\\
      `you! move up a little!'
\item \label{e05b} \mapuex{shinge \hspace{20pt} -nak \hspace{18pt} -üm \hspace{2pt} -ün} \hfill (de Augusta in \citealp{chandia2014})\\
      \gloss{IV.move-on AV.down +CA +PVN}\\
      `to pull something down, e.g. clothes when undressing'
\end{enumerate}
\normalsize

\cite{smeets2008}'s argument that these roots generate inflected forms is not consistently supported. For \mapu{shinge}, she provides only two examples in her entire work: one is (\hyperref[e05a]{E.05.a}) above; the other appears in the definition of \mapu{shinge} exclusively as an `auxiliary' (\mapu{shinge amunge} `move up!'). There is no dictionary entry for \mapu{shinge} as an independent verb root, nor any example of it hosting inflection. This absence is telling, as it suggests \mapu{shinge} may be the only genuine auxiliary in her set – a possibility we evaluate in \S\ref{sec.04.3}.

As defined above, and illustrated in (\hyperref[e01a]{E.01.a}) and (\hyperref[e04a]{E.04.a}), the ``auxiliary'' does not govern negation or inflection as \cite{moreno2000} would assert. \cite{bybee1994} also asserts that auxiliaries do not function as independent verbs, which the five auxiliaries listed by \cite{smeets2008} contradict in the (b) examples. Moreover, even in the (a) examples, these verbs retain lexical meaning, which they would not under \cite{bybee1994}'s terms. \cite{payne1997}'s criterion shows these elements do not meet phonological bounding with the main verb, as they appear both isolated and bound (see \hyperref[tab01]{Table 1}).

The distributional evidence strengthens the case against an auxiliary analysis. Consider the paradigm of imperative combinations with \mapu{kalli} across the corpus:

\begin{itemize}
\item 1sg imperative (\mapu{-chi}): \mapu{kalli müle-chi} `let me be' \citep{febres1846}
\item 3 imperative (\mapu{-pe}): \mapu{kalli amu-pe} `let him go' \citep{febres1846}
\item 2sg imperative (\mapu{-nge}): \textbf{attested nowhere in the corpus we examined}
\end{itemize}

A systematic search of the entire corpus—spanning four centuries of documentation—identified 39 attestations of \mapu{kalli} in imperative constructions. Of these, 30 (76.9\%) occur with the 3sg imperative suffix \mapu{-pe}, and 9 (23.1\%) with the 1sg imperative suffix \mapu{-chi}. Strikingly, not a single attestation occurs with any 2nd-person imperative form: neither 2sg (\mapu{-nge}), 2pl (\mapu{-m-ün}), nor 2dl (\mapu{-m-u}). This systematic gap is unlikely to be accidental.

Regarding the 2sg zero-marked form, a methodological caveat is in order. The 2sg intransitive imperative in \mapu{Mapudüngun} is formally zero-marked (bare stem), which makes it difficult to identify conclusively in the corpus. However, this does not weaken our argument. First, the transitive 2sg imperative is overtly marked with \mapu{-nge} (e.g., \mapu{pi-fi-nge} `say it! [to one person]'). Across our corpus of 39 imperative attestations with \mapu{kalli}, we found zero instances of \mapu{-nge}—a form that would be clearly identifiable if it occurred. Second, the 2pl and 2dl imperatives are also overtly marked: \mapu{-m-ün} and \mapu{-m-u} respectively. We found zero instances of either form. The restriction is thus not a consequence of zero-marking, but a systematic gap affecting all overtly marked 2nd-person forms.

If \mapu{kalli} were an independent auxiliary verb capable of hosting its own inflection, we would expect a full range of imperative forms. Its complete absence from all overtly marked 2nd-person constructions strongly suggests that \mapu{kalli} is not an independent verb in these contexts, but part of a complex predicate whose subject restrictions are determined by the final suffix.

\cite{smeets2008}'s analysis is pioneering, identifying \mapu{kalli, kim, küpa, pepi}, and \mapu{shinge} as auxiliaries based on their functional resemblance to cross-linguistic TAM markers. While her framework provides a valuable starting point, closer examination reveals complexities that invite alternative interpretations.

The observation that these elements occupy a fixed preverbal slot (e.g., \mapu{pepi küdaw-la-n}) aligns with auxiliary-like behaviour in other languages (cf. English \textit{will} + verb). They consistently encode modality (\mapu{küpa} `want', expressing volition, a sub-type of modality), aspect (\mapu{shinge} `move along', conveying progressive or directional aspect), or voice (\mapu{kalli} `let', functioning as a permissive, a type of causative voice), mirroring auxiliary roles across languages. In their preverbal use, they exhibit an absence of derivational morphology, supporting their categorisation as a distinct class.

Nevertheless, the data present ambiguities that complicate the auxiliary label. While \cite{smeets2008} acknowledges the capacity of these roots to function as independent verbs (e.g., \mapu{küpa-nu-l-i-iñ} `if we do not come'), this duality is atypical for auxiliaries, which characteristically undergo loss of lexical versatility \citep{bybee1994}. In the imperative \mapu{kalli düngu-pe} `let him speak!', the focus is on \mapu{düngu} rather than \mapu{kalli}, contrary to auxiliary behaviour where inflection typically attaches to the auxiliary \citep{moreno2000}. Native writers alternate between fused (\mapu{pepi-weyel-üy} `he can swim') and separated (\mapu{pepi weyelüy}) forms, suggesting perceptual ambiguity about their boundedness—possibly reflecting Spanish influence or Eurocentric linguistic descriptions.

The classification proposed by \cite{smeets2008} may reflect a stage of partial grammaticalisation, where these elements exhibit auxiliary-like properties without full convergence. For instance, \mapu{pepi} `can' retains its ability semantics while also serving a grammatical function—a situation distinct from fully bleached auxiliaries like English \textit{will} (which has lost its original volitional meaning). The elements' separability (\mapu{pepi weyel-i-ø} / \mapu{pepi-weyel-i-ø}) may signal this semi-grammaticalised status, with fusion representing the older lexical compound structure and separation reflecting the newer, Spanish-influenced analytic pattern.

These observations raise fundamental questions about how to categorise these elements. Four logical possibilities present themselves, each corresponding to a position in the literature:

\begin{itemize}
\item reclassify them as preverbal particles, following \cite{zuniga2006}'s descriptive practice;
\item regard them as V1 in verb-verb compounds, consistent with \cite{zuniga2006}'s radical concatenation model;
\item retain the auxiliary label while acknowledging their hybrid nature;
\item reclassify them as clitics—morphemes that are syntactically independent but phonologically dependent on a host word (a position implicit in \cite{zuniga2006}'s observation of gradient bonding).
\end{itemize}

The remainder of this paper argues for the second option—V1 in compounds—but not on syntactic grounds alone. Rather, we will show that the apparent evidence for the other possibilities arises from interpreting orthographic conventions as grammatical structure.

\begin{table}[H]
\centering
\caption{Diagnostic of \cite{smeets2008}'s Auxiliary Classification}
\label{tab01}
\vspace{0.2cm}
\begin{tabularx}{\linewidth}{lYYY}
\toprule
\textbf{Criterion} & \textbf{Expected Auxiliary Behaviour} & \textbf{\mapu{Mapudüngun} Pattern} & \textbf{Example} \\
\midrule
\textbf{Inflection Hosting} & Auxiliary bears inflection & Main verb hosts inflection & \mapu{pepi küdaw-\textbf{la}-n} (negation on \mapu{küdaw}) \\
\textbf{Syntactic Binding} & Phonologically bound to main verb & Variable separation/fusion & \mapu{pepi-weyel} (fused) vs. \mapu{pepi weyel} (separated) \\
\textbf{Semantic Bleaching} & Lexical meaning lost & Full retention of original meaning & \mapu{küpa} `want/come' \\
\textbf{Closed Class} & Fixed, small set & \mapu{shinge} excluded from verb list; class potentially open & Missing in \cite{smeets2008}'s dictionary as verb root \\
\textbf{Dependency} & Governs main verb syntax & Main verb governs syntax & \mapu{kalli düngu-\textbf{pe}} (imperative on \mapu{düngu}) \\
\bottomrule
\end{tabularx}
\end{table}

\vspace{0.5cm}
\subsubsection{\label{sec.01.1.2} Modal Prefixes \citep{longkon2011}}

According to \citeauthor{dixon2010} (\citeyear{dixon2010}: 92), modal prefixes are bound morphemes that attach to verb roots to express modality (necessity, possibility, or volition). These morphemes form a phonological unit with the verb and lack independent syntactic status (e.g., Ainu\footnote{Ainu is a language isolate traditionally spoken in Hokkaido (Japan) and Sakhalin (Russia). The prefix \textit{e-} marks obligative modality and attaches directly to the verb root, forming a single phonological word with no independent syntactic status \citep{dixon2010}. This exemplifies the typological expectation for modal prefixes—obligatory boundness—that the \mapu{Mapudüngun} elements do not satisfy.} \textit{e-} `must' in \textit{\textbf{e}-rap} `must kill').

\citeauthor{aikhenvald2006} (\citeyear{aikhenvald2006}: 36) specifies that prototypical modal prefixes exhibit three key properties: (1) fixed pre-root position, (2) obligatory attachment, and (3) semantic scope over the entire verb phrase. These elements typically constitute a closed class, as illustrated by Tariana \textit{-kade-} `be able to'.

\citeauthor{payne1997} (\citeyear{payne1997}: 244) concurs, emphasising that modal prefixes are phonologically bound and strictly precede the root without intervening material. Their functional scope is restricted to marking TAM.

\citeauthor{longkon2011} (\citeyear{longkon2011}: 249) designates the elements under discussion as `modal prefixes', identifying four: \mapu{kalli, kim-, küpa-} and \mapu{pepi-}, assigning semantic values aligning with \cite{smeets2008}. Her framework posits these as prefixes, grounded on the premise that they are invariably attached to the verb they influence, except for the permissive \mapu{kalli}, which never forms part of a compounded form when functioning as a modal prefix. Examples (\hyperref[e06]{E.06}) illustrate this distinction:

\paragraph{\example{06} \label{e06}}~
\vspace{-10pt}\begin{enumerate}[label=\alph*.]
\item \label{e06a} \mapuex{kalli \hspace{38pt} düngu \hspace{12pt} -pe}\\
      \gloss{MP.enabling IV.speak +IMP.3}\\
      `let that he speaks (let him speak)'
\item \label{e06b} \mapuex{kim \hspace{20pt} -rüngkü \hspace{6pt} -y \hspace{8pt} -ø}\\
      \gloss{MP.know IV.jump +IND +3}\\
      `she knows [how to] jump'
\item \label{e06c} \mapuex{küpa \hspace{16pt} -yi \hspace{16pt} -ñ}\\
      \gloss{MP.wish TV.eat +IND.1SG}\\
      `I want to eat'
\item \label{e06d} \mapuex{pepi \hspace{14pt} -neng \hspace{10pt} -üm \hspace{2pt} -üy \hspace{4pt} -ø}\\
      \gloss{MP.can IV.move +CA +IND +3}\\
      `she can move [it]'
\end{enumerate}
\normalsize

As in (\hyperref[e01a]{E.01.a}), the imperative affix in (\hyperref[e06a]{E.06.a}) attaches not to the modal prefix \mapu{kalli} `let', but to the bare infinitive (verbalized noun) \mapu{düngu} `to speak'. As with \cite{smeets2008}, this appears to contradict \cite{longkon2011}'s proposal, as the translation suggests the exhortation is directed toward the second person, while the imperative morpheme relates to the third person executing the action. By this reasoning, \mapu{kalli} should be bound, forming a verb stem with \mapu{düngu}.

The ambiguity of (\hyperref[e06a]{E.06.a}) is instructive. A literal reading of the written form – \mapu{kalli} as a separate word followed by the imperative \mapu{düngu-pe} – would yield a semantically contradictory command: a 3rd person imperative (`let that he speak!') cannot simultaneously be a 2nd person command to `let'. The only coherent interpretation requires treating \mapu{kalli} and \mapu{düngu} as a single unit over which the imperative scopes: \mapu{kalli-düngu-pe}. The space in the written form is thus misleading; it reflects orthographic convention, not grammatical structure.

From the definitions above, a contradiction emerges between \cite{longkon2011}'s characterisation and established prefix criteria. The forms exhibit variable binding, whereas \cite{dixon2010} asserts that prefixes must be obligatorily bound. This discrepancy is exemplified by \mapu{kalli}, depicted as a distinct word in imperatives (\mapu{kalli düngu-pe!}).

Regarding semantic scope, \cite{aikhenvald2006} expects such prefixes to modify the entire verb phrase. However, as (\hyperref[e06a]{E.06.a}) demonstrates, the imperative \mapu{-pe} scopes only over \mapu{düngu-} `speak', while \mapu{kalli} remains outside its domain.

Concerning phonological unity, \cite{payne1997} states that prefixes should not permit orthographic separation. Nevertheless, native speaker orthographic conventions (e.g., writing \mapu{kalli} separately) contradict this principle, as does the variable binding of the same form.

While \cite{longkon2011}'s classification captures certain syntactic and semantic patterns – fixed preverbal position (e.g., \mapu{küpa-yi-ñ} `I want to eat', where \mapu{\textbf{*yi-ñ küpa}} is ungrammatical) and modal semantics (\mapu{pepi-} `ability') – the analysis is problematic for two reasons. First, it fails to account for variable binding properties, violating \cite{dixon2010}'s criterion of obligatory attachment for true prefixes. Second, it overlooks dual lexical-grammatical function, as these forms retain independent verbal uses outside modal constructions.

This oversimplification becomes apparent when considering \cite{zuniga2006}'s observation of gradient bonding, suggesting clitic-like behaviour rather than prototypical prefixation. Furthermore, their retained verbal properties align more closely with V1 elements in compound structures than with inflectional prefixes. Consequently, while the prefix label may account for positional regularity, it ultimately obscures the grammatical complexity of these elements.

Further complicating \cite{longkon2011}'s analysis are examples from her own data that contradict the claim of obligatory prefixation. For \mapu{kim}, she provides (2b) \mapu{kim düngu-y ti pichi elaweni} `that child knows how to speak' (\citeyear{longkon2011}: 208) – a clear instance of separation. For \mapu{pepi}, example (73) \mapu{tüfachi ruka pepi müle-nge-la-y} `this house cannot be inhabited' (\citeyear{longkon2011}: 250) shows the same pattern. These are not exceptions; they are the norm that her prefix analysis cannot accommodate.

The negation data provide an even stronger diagnostic. In every attested example – from \cite{febres1765} to \cite{conadi1996b} – negation attaches to the second element, never to \mapu{kalli} itself:

\paragraph{\example{07} \label{e07}}~
\vspace{-10pt}\begin{enumerate}[label=\alph*.]
\item \label{e07a} \mapuex{kalli \hspace{8pt} mupi \hspace{24pt} -no \hspace{8pt} -a \hspace{10pt} -l \hspace{10pt} -m \hspace{2pt} -i} \hfill \citep{febres1846}\\
      \gloss{let AJ.truthful +NEG +FUT +SBJ +2 +SG}\\
      `provided you don't tell the truth'
\item \label{e07b} \mapuex{kalli \hspace{8pt} pe \hspace{8pt} -no \hspace{8pt} -l \hspace{14pt} -i \hspace{4pt} -ø} \hfill  \citep{febres1846}\\
      \gloss{let TV.see +NEG +SBJ +1 +SG}\\
      `when/if I don't see it'
\item \label{e07c} \mapuex{kalli \hspace{8pt} pe \hspace{10pt} -no \hspace{8pt} -lu \hspace{8pt} iñche} \hfill \citep{febres1846}\\
      \gloss{let tv.see +NEG +SVN +PP.I}\\
      `if I don't see it'
\item \label{e07d} \mapuex{kalli \hspace{8pt} ta \hspace{20pt} -ñi \hspace{10pt} pe \hspace{14pt} -no \hspace{8pt} -el} \hfill \citep{febres1846}\\
      \gloss{let AP.the +SP.my TV.see +NEG +OVN}\\
      `without my seeing it'
\item \label{e07e} \mapuex{kalli \hspace{6pt} ngüne \hspace{16pt} -w \hspace{8pt} -ki \hspace{10pt} -l \hspace{10pt} -m \hspace{4pt} -i}\footnote{The verb form in (\hyperref[e07d]{E.07.d}) integrates the negative suffixes for imperatives (\mapu{-ki-l-}) to an indicative form, the imperative form for a 2nd person should be: \mapu{kalli ngüne-w-ki-l-nge}.} \hfill \citep{febres1846}\\
      \gloss{let TV.govern +REF +NEG +SBJ +2 +SG}\\
      `don't govern yourself'
\end{enumerate}
\normalsize

Yet in every case, the negation semantically scopes over the entire `letting + action' complex. This is exactly what we expect if \mapu{kalli} and the following verb form a single grammatical unit; it is inexplicable if \mapu{kalli} is an independent prefix or auxiliary or particle.

\subsubsection{\label{sec.01.1.3} Preverbal Particles and Complex Verb Stems \citep{zuniga2006}}

Theoretical foundations for these concepts are provided by \citeauthor{payne1997} (\citeyear{payne1997}: 211), who defines preverbal particles as free morphemes occupying fixed preverbal position to encode TAM. These elements are phonologically independent but syntactically bound to the verb phrase (e.g., English \textit{must} in \textit{must go}). \citeauthor{aikhenvald2006} (\citeyear{aikhenvald2006}: 58) characterises these particles as non-fused modifiers, sharing functional scope with affixes but being separable, in contrast to true prefixes (e.g., Tariana \emph{kade} as particle vs. \emph{-kade-} as prefix).

Complex verb stems (CVS) are formed through compounding of verb roots, where one root modifies another semantically or grammatically, forming a single phonological and syntactic unit (\citealp{dixon2010}: 175). Yimas illustrates this: \emph{kpa-ŋka-} `want-eat'. \citeauthor{baker1996} (\citeyear{baker1996}: 320) further specifies that CVS in polysynthetic languages often involve noun incorporation or radical concatenation, with the first verb (V1) modifying the second (V2).

\citeauthor{zuniga2006} (\citeyear{zuniga2006}: 136) identifies two modal elements in \mapu{Mapudüngun – kim-} `know how' and \mapu{pepi-} `can' – forming part of what he terms complex verb stems (CVS), where both roots form a compound. However, he also posits these elements can be expressed as separate entities, designated preverbal particles (PVP). Examples (\hyperref[e08]{E.08})–(\hyperref[e09]{E.09}) illustrate:

\paragraph{\example{08} \label{e08}}~
\vspace{-10pt}\begin{enumerate}[label=\alph*.]
\item \label{e08a} \mapuex{kim \hspace{48pt} weyel \hspace{10pt} -i \hspace{10pt} -ø}\\ 
      \gloss{PVP.know-how IV.swim +IND +3}\\
      `he can swim / he knows how to swim'
\item \label{e08b} \mapuex{kim \hspace{38pt} -weyel \hspace{8pt} -i \hspace{12pt} -ø}\\
      \gloss{AJ.knower +IV.swim +IND +3 (CVS)}\\
      `he can swim / he knows how to swim'
\end{enumerate}
\normalsize

\paragraph{\example{09} \label{e09}}~
\vspace{-10pt}\begin{enumerate}[label=\alph*.]
\item \label{e09a} \mapuex{pepi \hspace{36pt} weyel \hspace{10pt} -i \hspace{10pt} -ø}\\ 
      \gloss{PVP.be-able IV.swim +IND +3}\\
      `he can swim / he is enabled to swim'
\item \label{e09b} \mapuex{pepi \hspace{20pt} -weyel \hspace{8pt} -i \hspace{10pt} -ø}\\
      \gloss{TV.can +IV.swim +IND +3 (CVS)}\\
      `he can swim / he is enabled to swim'
\end{enumerate}
\normalsize

\citeauthor{zuniga2006} (\citeyear{zuniga2006}: 179) presents a list of verb roots forming complex verb stems, including \mapu{wüño-} `re-' ← `return, come back' – the focus of this article (see \S\ref{sec.01.1.4}). The list also includes \mapu{kalli-} `enable, let' and \mapu{küpa-} `wish', but not \mapu{shinge-}, listed by \cite{smeets2008}.

\cite{zuniga2006} explains that these forms frequently appear as preverbal particles – separated from the main verb, reflected in spelling – yet he treats them as radical concatenation.

According to  \cite{payne1997}'s definition, preverbal particles must demonstrate phonological independence. In contrast, \cite{zuniga2006}'s CVS analysis conflates fused (prefix-like) and separated (particle-like) forms, categorising them collectively.

Conversely, Complex Verb Stems are expected to manifest unitary behaviour \citep{dixon2010}. However, \mapu{kim-} and \mapu{pepi-} maintain distinct meanings outside the CVS framework (e.g., \mapu{kim} as stand-alone adjective `wise, the knower'). Orthographic inconsistency in native spelling (e.g., \mapu{pepi weyel} vs. \mapu{pepi-weyel}) suggests speakers perceive separability, undermining the CVS claim.

\cite{zuniga2006}'s proposal, however, provides a more illuminating view of elements seen as auxiliaries \citep{smeets2008} and modal prefixes \citep{longkon2011}. \cite{zuniga2006} highlights positional regularity – both PVPs and CVS occupy preverbal slots – and semantic transparency: \mapu{pepi-} consistently marks ability across fused/separated forms. But gradient binding cannot be ignored; alternation between PVPs and CVS aligns more with clitics (\citealp{zuniga2006}: 136) than rigid categories. Lexical retention respects independent uses (\mapu{küpa} `wish' as full verb), contradicting bound-stem assumptions of CVS (V1 modifies V2).

\cite{zuniga2006}'s framework usefully captures the continuum between fusion and separation, but overstretches the CVS label. A hybrid analysis – treating these elements as clitic-like modifiers – better accounts for their dual behaviour (cf. \citealp{aikhenvald2006} on Tariana).

\subsubsection{\label{sec.01.1.4} Modals, Prefixes, and the Status of \mapu{wüño}}

This section builds on the terminological framework established in \S\ref{sec.01.1.2} (\citealp{longkon2011}'s modal prefixes) and \S\ref{sec.01.1.3} (\citealp{zuniga2006}'s preverbal particles), examining \mapu{wüño}'s dual behaviour as both a modal element and a potential prefix.

\citeauthor{salas1992a} (\citeyear{salas1992a}: 192) classifies the preverbal elements discussed here as modals, identifying \emph{kim-} `to know', \mapu{küpa-} `to wish' and \mapu{pepi-} `to be able'. He argues they function as prefixes attached to simple and complex stems, aligning with \cite{longkon2011}'s proposal (see \S\ref{sec.01.1.2}).

de Augusta defines \mapu{wüño-} as a prefix equivalent to Spanish \emph{re-}, expressing repetition of the action indicated by the following root -- during the research, \mapu{wüño-} has also been found to express retrospective and/or backward action. Among examples in de Augusta's dictionary \citep{chandia2014}, \mapu{wüño-} appears both separated from the main verb and compounded with the next root.

Occurrences of \mapu{wüño-} give the impression that moulding prevails over derivation from interaction of the two verb roots. If moulding prevails, \mapu{wüño-} could function as a modal verb, as in these examples from de Augusta's dictionary:

\paragraph{\example{10} \label{e10}}~
\vspace{-10pt}\begin{enumerate}[label=\alph*.]
\item \label{e10a} \mapuex{wüño \hspace{10pt} fey \hspace{22pt} -pi \hspace{14pt} -n}\\ 
      \gloss{MV.re DP.that +TV.say +IND.1SG}\\
      `answer' (lit.: `return that say')
\item \label{e10b} \mapuex{wüño \hspace{44pt} -kintu \hspace{26pt} -n}\\
      \gloss{IV.turn-back +TV.look-for +PVN}\\
      `to look back'
\item \label{e10c} \mapuex{wüño \hspace{38pt} -kutran \hspace{12pt} -tu \hspace{8pt} -n}\\
      \gloss{IV.turn-back NN.illness +APL +PVN}\\
      `to have a relapse' (lit.: `return ill be')
\item \label{e10d} \mapuex{wüño \hspace{12pt} tu \hspace{16pt} -n}\\
      \gloss{MV.re TV.take +PVN}\\
      `recover possession of something' (lit.: `return take')
\end{enumerate}
\normalsize

In these examples, the main verb retains its meaning while \mapu{wüño} adds the idea of doing back (\hyperref[e10a]{E.10.a}, \hyperref[e10b]{E.10.b}) or recovering (\hyperref[e10c]{E.10.c}, \hyperref[e10d]{E.10.d}).

Many native speakers perceive \mapu{wüño} as separate from the verb it supports, reflecting it in spelling. Examples (\hyperref[e11]{E.11}) demonstrate this:

\paragraph{\example{11} \label{e11}}
\vspace{-10pt}\begin{enumerate}[label=\alph*.]
\item \label{e11a} \mapuex{wüño \hspace{3pt} waychüf \hspace{3pt} -küno \hspace{10pt} -w \hspace{10pt} -a \hspace{12pt} -y \hspace{10pt} -ø \hspace{2pt} antü} \hfill (Lienlaf, T. in \citealp{aldunate-lienlaf2002})\\
      \gloss{MV.re IV.roll +TV.let +REF +FUT +IND +3 NN.day}\\
      `they will be going around [all] day'
\item \label{e11b} \mapuex{ka \hspace{18pt} wüño \hspace{10pt} aku \hspace{18pt} -tu \hspace{6pt} -ke \hspace{10pt} -y \hspace{10pt} -ø \hspace{4pt} alü \hspace{16pt} -n} \hfill (Ayllapan, L. in \citealp{aldunate-lienlaf2002})\\
      \gloss{CJ.and MV.re IV.arrive +APL +HAB +IND +3 AJ.very +PVN}\\
      `and again he is coming away [as he always does]'
\item \label{e11c} \mapuex{tüfa \hspace{18pt} ka \hspace{14pt} wüño \hspace{8pt} tu \hspace{16pt} -wü \hspace{2pt} -la \hspace{10pt} -n \hspace{24pt} ñi \hspace{16pt} ketra \hspace{20pt} -n} \hfill \citep{wirimilla2005b}\\
      \gloss{DP.now CJ.too MV.re TV.take +PS +NEG +IND.1SG SP.my TV.plough +PVN}\\
      `now I have not taken up my ploughing again either'
\item \label{e11d} \mapuex{wüño \hspace{8pt} treka \hspace{12pt} -t \hspace{10pt} -i \hspace{12pt} -ø \hspace{6pt} pun} \hfill (\citealp{marileo2006}: 18)\\
      \gloss{MV.re IV.walk +APL +IND +3 NN.night}\\
      `the night walks back'
\item \label{e11e} \mapuex{welu \hspace{10pt} wüño \hspace{10pt} ina \hspace{12pt} -tu \hspace{8pt} -fi \hspace{8pt} -l \hspace{12pt} -e \hspace{8pt} düngu \hspace{12pt} engün \hspace{10pt} lleg \hspace{24pt} -pu \hspace{22pt} -a \hspace{10pt} -fu \hspace{2pt} -y}\\
      \gloss{CJ.but MV.re AV.next +APL +3P +SBJ +3 NN.matter PP.they IV.grow +IV.arrive +FUT +RI +3}\\
      `but the matter is again explained to them until they arrive to grow' \hfill \citep{marimanetal2006}
\item \label{e11f} \mapuex{fey \hspace{28pt} -ti \hspace{22pt} -chi \hspace{4pt} wiño \hspace{8pt} ye \hspace{24pt} -ñpüra \hspace{14pt} -m \hspace{8pt} -nge \hspace{6pt} -tu \hspace{10pt} -n} \hfill \citep{indh2011}\\
      \gloss{DP.this +AP.that +ADJ MV.re TV.carry +IV.go-up +CA +IV.be +APL +PVN}\\
      `those [who] rise again'
\item \label{e11g} \mapuex{ta \hspace{24pt} ñi \hspace{20pt} wiño \hspace{10pt} nü \hspace{16pt} -tu \hspace{10pt} -a \hspace{10pt} -fi \hspace{8pt} -el} \hfill \citep{kidel2006} \\
      \gloss{AP.the SP.their MV.re TV.take +APL +FUT +3P +OVN}\\
      `they will take them again'
\item \label{e11h} \mapuex{ñi \hspace{18pt} wiño \hspace{10pt} nentu \hspace{22pt} -w \hspace{10pt} -tu \hspace{10pt} -a \hspace{10pt} -f \hspace{6pt} -el} \hfill (Alvear, G. in \citealp{mineduc2005ad})\\
      \gloss{SP.his MV.re TV.take-out +REF +APL +FUT +RI +OVN}\\
      `his repeated efforts' (lit.: his repeated taking out of himself)
\item \label{e11i} \mapuex{ñi \hspace{16pt} wiño \hspace{4pt} wütra \hspace{6pt} -m \hspace{6pt} -püra \hspace{10pt} -m \hspace{8pt} -nge \hspace{8pt} -tu \hspace{8pt} -a \hspace{12pt} -l \hspace{16pt} ñi} \hfill \citep{zunigaolate2017} \\
      \gloss{SP.its MV.re IV.rise +CA IV.go-up +CA +IV.be +APL +FUT +OVN SP.its}\\
      \mapuex{matu \hspace{18pt} -matu \hspace{16pt} wiño \hspace{8pt} yafü \hspace{12pt} -l \hspace{10pt} -üw \hspace{8pt} -a \hspace{10pt} -l}\\
      \gloss{AJ.quick +AJ.quick MV.re AJ.hard +CA +REF +FUT +OVN}\\
      `its effective and efficient revitalization in the short term' (lit.: its being risen up again [and] its quick hardening again)
\item \label{e11j} \mapuex{ka \hspace{20pt} chum \hspace{10pt} -nge \hspace{8pt} -ø \hspace{12pt} -chi \hspace{4pt} wiño \hspace{3pt} wütra \hspace{5pt} -m \hspace{6pt} -püra \hspace{12pt} -m \hspace{6pt} -nge \hspace{8pt} -tu \hspace{8pt} -a \hspace{10pt} -fu \hspace{4pt} -y \hspace{8pt} -ø}\\
      \gloss{CJ.and QC.how +IV.be +SVN +ADJ MV.re IV.rise +CA IV.go-up +CA +IV.be +APL +FUT +RI +IND +3}\\
      `in its maintenance and development' (lit.: and how it will be risen up again) \hfill \citep{longkon2017}
\end{enumerate}
\normalsize

De Augusta states that \mapu{wüño} functions as a prefix. However, no affix in \mapu{Mapudüngun} occurs in isolation; it must attach to a verb or other root. Additionally, an affix cannot function as a root that accepts suffixes. In contrast, \mapu{wüño} operates as a verb root that can be inflected by attaching suffixes (see \hyperref[e12]{E.12}).

\paragraph{\example{12} \label{e12}}
\vspace{-10pt}\begin{enumerate}[label=\alph*.]
\vspace{-27pt}
\item[] \mapuex{wüño \hspace{22pt} -me \hspace{4pt}  -n}  \hfill  (de Augusta in \citealp{chandia2014})\\
\hspace{24pt}  \gloss{IV.return +TH +IND.1SG}\\
\hspace{24pt} `I went back there (to that place)'
\end{enumerate}
\normalsize

\mapu{Wüño} either forms a compound or appears as a separate element. If some native speakers perceive it as separate, this may indicate it functions as a modal. Following \cite{smeets2008}, this verb would fulfil the role of an `auxiliary', as she defines it, or a `preverbal particle', as \citeauthor{zuniga2006} (\citeyear{zuniga2006}: 136) describes it. \citeauthor{smeets2008} (\citeyear{smeets2008}: 175) states that `an auxiliary is an uninflected verb stem which immediately precedes (there are no elements allowed in between) the verb with which it is combined'.

Not all native speakers spell \mapu{wüño} separately. Many form a verb stem by compounding, treating it as a `modal prefix' (\citealp{longkon2011}: 249). However, in some cases (\hyperref[e01a]{E.01.a}, \hyperref[e04a]{E.04.a}, \hyperref[e06a]{E.06.a}), usage appears inconsistent, possibly due to influence from standard uses of `auxiliaries' or Spanish grammar, which presents modal verbs as separate elements.

In summary, \mapu{wüño} fails to meet the criterion of phonological autonomy expected of prefixes \citep{dixon2010}, which are obligatorily bound morphemes. Instead, it exhibits variable binding, appearing both fused (\mapu{wüño-kintu-n} `to look back') and separated (\mapu{wüño tu-n} `recover'). This variation is reflected in native orthographic practices (\hyperref[e11a]{E.11.a}–\hyperref[e11j]{E.11.j}), suggesting speakers perceive it as a distinct element.

Morphosyntactically, \mapu{wüño} behaves as a verb root rather than an affix: unlike prefixes, it can host suffixes (e.g., \mapu{wüño-me-n} `I returned [there]', \hyperref[e12]{E.12}). This aligns with \cite{smeets2008}'s definition of auxiliaries as uninflected stems preceding main verbs (e.g., \mapu{wüño waychüf-küno-w-a-y-ø}, \hyperref[e11a]{E.11.a}), while its semantic role – adding repetition or retrospection without altering the main verb's core meaning – parallels \cite{zuniga2006}'s preverbal particles (cf. \S\ref{sec.01.1.3}).\\

Reconciling the evidence:
\begin{itemize}
\item \textbf{Against prefix status:}
    \begin{itemize}
    \item Fails binding and inflection tests (\S\ref{sec.01.1.2})
    \item Orthographic variation (\mapu{wüño} bound vs. unbound) suggests native speakers reject full fusion\\
    \end{itemize}
\item \textbf{For modal/auxiliary status:}
    \begin{itemize}
    \item Displays positional regularity (preverbal) and semantic transparency (\S\ref{sec.01.1.3})
    \item Retains verbal properties (e.g., \mapu{wüño-me-n} `I returned [there]')
    \end{itemize}
\end{itemize}

Typologically, \mapu{wüño}'s clitic-like gradience mirrors Tariana's \textit{-kade} \citep{aikhenvald2006}, bridging prefixes and particles. While \mapu{wüño} exhibits prefix-like fusion in some contexts, its morphosyntactic flexibility and orthographic treatment support reclassifying it as a clitic or preverbal modifier, reconciling de Augusta's prefix label with contemporary usage.

Thus, \mapu{wüño} epitomises the central analytical dilemma presented by all preverbal elements under review: it displays characteristics of prefixes, particles, and independent verbs, yet fits neatly into none of these categories. Its dual behaviour mirrors contradictions observed with \mapu{kalli, kim, küpa, pepi}, and \mapu{shinge} – contradictions that have generated the competing classifications surveyed in \S\ref{sec.01.1.1}–\S\ref{sec.01.1.3}. This irreducible hybridity suggests that the very framework of analysis, which seeks to assign discrete grammatical labels based primarily on orthographic spacing, may be fundamentally misaligned with the grammatical reality of \mapu{Mapudüngun}.

\subsubsection{\label{sec.01.1.5} Summary: The Analytical Dilemma}

The preceding review reveals a fundamental problem in the analysis of \mapu{Mapudüngun}'s preverbal elements. Each scholarly framework—auxiliaries \citep{smeets2008}, modal prefixes \citep{longkon2011}, and preverbal particles/complex verb stems \citep{zuniga2006}—captures genuine features of these elements' behaviour, yet each encounters irreducible contradictions when confronted with the full range of empirical evidence.

The central dilemma can now be stated precisely: the preverbal elements appear in constructions that have been analysed as auxiliaries, prefixes, particles, or full lexical verbs, depending on the framework applied. This variability suggests that the elements themselves may not belong to a single discrete category, but rather occupy different points on a continuum depending on the construction and the diagnostic criteria used. The orthographic record, meanwhile, presents a confusing picture of alternating fusion and separation that correlates with neither grammatical function nor semantic change.

This situation suggests that the problem may lie not in the elements themselves, but in the interpretive framework applied to them. All existing analyses share a crucial assumption: that the orthographic forms preserved in written sources directly reflect grammatical structure. If this assumption is false—if the written record is shaped by factors independent of grammar—then the apparent contradictions may dissolve.

The following section \ref{sec.02} develops this insight into a positive proposal: the \textbf{prosodic-orthographic hypothesis}. We argue that the variable binding recorded in the corpus may result from the fossilisation of prosodic pauses transcribed by early missionaries as spaces, a convention subsequently inherited and reinterpreted by later writers and scholars. On this view, the underlying grammar of \mapu{Mapudüngun} is, and always has been, \cite{zuniga2006}'s ``radical concatenation'', with these elements functioning as ordinary verb roots in V1 position of compounds. The illusion of `hybrid' behaviour would then arise from interpreting orthographic conventions as syntactic evidence.

\section{\label{sec.02} The Prosodic-Orthographic Hypothesis}

The preceding review (\S\ref{sec.01}) has identified a fundamental analytical dilemma: the preverbal elements appear in constructions that have been analysed as auxiliaries, prefixes, particles, or full lexical verbs, depending on the framework applied (\S\ref{sec.01.1.1}–\S\ref{sec.01.1.4}). This variability suggests that the elements themselves may not belong to a single discrete category, but rather occupy different points on a continuum depending on the construction and the diagnostic criteria used.

Crucially, all previous analyses share a common assumption: that the orthographic forms preserved in written sources directly reflect grammatical structure. If this assumption is false—if the written record is shaped by factors independent of grammar—then the apparent contradictions may dissolve.

This section develops that insight into a positive proposal: the \textbf{prosodic-orthographic hypothesis}. We argue that the variable binding observed in the corpus may result from the fossilisation of prosodic pauses transcribed by early missionaries as spaces, a convention subsequently inherited and reinterpreted by later writers and scholars. On this view, the underlying grammar of \mapu{Mapudüngun} is, and always has been, \cite{zuniga2006}'s radical concatenation, with these elements functioning as ordinary verb roots in V1 position of compounds. The illusion of hybrid behaviour would then arise from interpreting orthographic conventions as syntactic evidence.

\subsection{\label{sec.02.1} The Prosodic-Structural Continuum}

The variation observed in the corpus can be modelled as a five-stage historical continuum, each stage representing a different layer in the sedimentation of orthographic practice atop grammatical structure.

\subsubsection{\label{sec.02.1.1} Stage 1: Original Grammatical Structure (pre-1606)}

Before European contact, \mapu{Mapudüngun} exhibited true \textbf{radical concatenation} \citep{zuniga2006}, forming V1–V2 compounds where the first element modified the second. In this oral-only stage, no orthography obscured the grammatical reality: these elements were simply verb roots occupying the first slot in a productive compounding system, no different from any other verb root in the language. Crucially, as we will see in \S\ref{sec.02.2.4}, the system was category-neutral and recursive, admitting nouns, adjectives, and adverbs alongside verbs as modifiers—a pattern that confirms the preverbal elements are not a special class, but ordinary participants in a general grammatical process.

\subsubsection{\label{sec.02.1.2} Stage 2: Early Transcription (1606–1765)}

Missionary linguists—de Valdivia (1606), Febrés (1765)—lacking a native written tradition to guide them, transcribed what they heard. Crucially, they heard not only the segmental content of utterances but also \textbf{prosodic features}: phrasal rhythm, emphasis, and notably, the final-syllable stress characteristic of \mapu{Mapudüngun} bisyllabic words (e.g., \mapu{wüñÓ}). \citeauthor{smeets2008} herself (\citeyear{smeets2008}: 176) provides the phonetic key: `Between \mapu{kim} and \mapu{tuku-fi-n} in (93) a pause can be heard which is lacking in (96).' These prosodic pauses, which reflected performance rather than grammar, were interpreted as word boundaries and transcribed as spaces. The result was an orthography that systematically overdifferentiated the grammar, writing as separate what was grammatically unitary.

As \citeauthor{deaugusta1934} (\citeyear{deaugusta1934}: VI) observe, compound verbs have two stress peaks, and separating the auxiliary from the root would facilitate reading. In his own words:

\begin{quote}
\mapu{Los verbos compuestos tienen dos acentos, uno principal en el primer elemento de la combinación, y otro secundario en la terminación... Mucho se facilitaría la lectura, si se separara del auxiliar \mapu{ngen} la raíz verbal.}
\end{quote}

This observation—that compound verbs have two stress peaks, and that separating the auxiliary from the root would facilitate reading—confirms that the spaces in missionary orthography reflect prosodic structure (specifically, the presence of two stress peaks in compound verbs) rather than syntactic independence.

\subsubsection{\label{sec.02.1.3} Stage 3: Orthographic Fossilisation (1765–1916)}

These missionary-created spaces became \textbf{fixed conventions} in subsequent grammars and dictionaries. \citeauthor{febres1765}' influential \textit{Arte de la Lengua General del Reyno de Chile} (\citeyear{febres1765}) established spacing patterns that later works—including Augusta's dictionary (1916)—inherited and reinforced. What began as a transcription artifact hardened into orthographic fact, creating the impression of syntactic independence where none existed in the spoken language. The spaces were now part of the textual tradition, copied from one source to the next without reference to the spoken forms that had originally motivated them.

\subsubsection{\label{sec.02.1.4} Stage 4: Scholarly Perpetuation (1916–1992)}

Linguists working from these written sources—\cite{smeets2008}, \cite{longkon2011}, and even \cite{zuniga2006} in his descriptive moments—interpreted the inherited orthography as direct evidence for grammatical status. The spaces in texts became the basis for claims about auxiliaries, prefixes, and particles, perpetuating the assumption that the missionary transcription reflected \mapu{Mapudüngun} syntax rather than European perceptual habits. The grammatical consensus of the late twentieth century was built on an orthographic foundation laid three centuries earlier by men who had never heard the language spoken in its pre-contact form.

\subsubsection{\label{sec.02.1.5} Stage 5: Native Speaker Reanalysis (1992–present)}

Contemporary native writers, educated in the written tradition, have internalised the orthographic conventions to the point of \textbf{syntactic reanalysis}. The spaces first introduced by missionaries are now treated as genuine grammatical boundaries, leading to innovations impossible in the original oral language—most notably, the insertion of adverbs and other material between the preverbal element and the main verb (see \S\ref{sec.04.5}). This final stage represents a true contact-induced change: not Spanish influence on \mapu{Mapudüngun} grammar directly, but Spanish-influenced orthography becoming grammar through four centuries of written practice. The insertion evidence in (\hyperref[e16]{E.16}–\hyperref[e18]{E.18}) documents this reanalysis in progress.

This five-stage model reframes the variable binding observed in the corpus. It is not free alternation between syntactic categories, but \textbf{gradient orthographic representation of a consistently compounded structure}, overlaid with later syntactic reanalysis in the most recent period. The fusion vs. separation seen in the corpus thus reflects not a grammatical switch, but the intersection of three distinct factors: original compounding structure, missionary transcription conventions, and contemporary reanalysis.

\subsection{\label{sec.02.2} Resolving the Contradictions}

The prosodic-orthographic hypothesis resolves the specific contradictions that undermined previous analyses. Each of the following subsections takes a diagnostic that proved problematic for earlier frameworks—prefixhood, auxiliary/particle status, the \mapu{küpa} semantics, category neutrality, and Spanish contact—and shows how the hypothesis transforms these apparent problems into confirming evidence.

\subsubsection{\label{sec.02.2.1} Against Pure Prefix Status}

The prefix analysis \citep{longkon2011} encounters two fundamental problems. First, true prefixes are obligatorily bound \citep{dixon2010}; they cannot appear as independent phonological words. Yet \mapu{pepi} is frequently written separately, and \mapu{wüño} hosts suffixes (\mapu{wüño-me-n} `I returned there', \hyperref[e12]{E.12}). Second, prefixes cannot occupy the second position in a compound. As we will see in \S\ref{sec.04.2}, the corpus documents V2 attestations for \mapu{kim} and \mapu{küpa}—forms where the supposed prefix is the incorporated element, not the incorporator. These patterns are not exceptions; they are the norm that the prefix analysis cannot accommodate.

Under the prosodic-orthographic hypothesis, these facts are unproblematic. \mapu{Kim}, \mapu{küpa}, and \mapu{pepi} are verb roots, fully capable of inflection when used independently and of occupying any position in a compound. The orthographic separation that sometimes appears is a later artifact of missionary transcription conventions, not a reflection of grammatical boundedness.

\subsubsection{\label{sec.02.2.2} Against Pure Auxiliary or Particle Status}

Cross-linguistically, true auxiliaries host inflection while the main verb remains invariant \citep{moreno2000}. Particles, by contrast, stand entirely outside the scope of verbal inflection \citep{payne1997}. The \mapu{Mapudüngun} elements fit neither profile.

Consider the imperative in \mapu{kalli düngu-pe} (\hyperref[e01a]{E.01.a}, \hyperref[e06]{E.06.a}). The suffix \mapu{-pe} attaches to the second element \mapu{düngu}, yet semantically scopes over the entire `let speak' complex. This is exactly what we expect if the underlying structure is a compound \mapu{kalli-düngu-pe}; a true auxiliary would require the inflection on itself, while a true particle would be inaccessible to suffixal scope altogether. The negation evidence is even stronger: in every attested example from \cite{febres1846} to \cite{conadi1996b}, negation attaches to the second element, never to \mapu{kalli} itself (\hyperref[e07a]{E.07.a}–\hyperref[e07e]{E.07.e}). Yet the negation semantically scopes over the entire `letting + action' complex—again, the hallmark of a unitary compound.

The ambiguity of the written form—\mapu{kalli} separated by a space—has misled previous analysts into seeing syntactic independence where none exists. The coherent interpretation of the meaning requires treating the sequence as a single grammatical unit. The spaces reflect prosodic transcription, not syntactic structure.

\subsubsection{\label{sec.02.2.3} The \mapu{küpa} `Come/Want' Problem Revisited}

The Eurocentric interpretation of \mapu{küpa} as a volitional modal ('want') rather than a motion verb ('come') illustrates the broader issue. Constructions like \mapu{küpa langümfiñ} (\hyperref[e13]{E.13}), traditionally glossed as `I want to kill him', are more literally `I come to kill him'—a motion construction interpreted through Spanish modal categories. \cite{guevara1913} documents this pattern explicitly: Spanish translators routinely rendered motion constructions as expressions of volition, projecting the Spanish modal frame onto a \mapu{Mapudüngun} motion structure.

\paragraph{\example{13} \label{e13}}~
\vspace{-10pt}\begin{enumerate}[label=\alph*.]
\vspace{-17pt}
\item [] \mapuex{küpa \hspace{4pt} lang \hspace{2pt} -üm \hspace{2pt} -fi \hspace{10pt} -ñ} \hfill \citep{guevara1913}\\
\gloss{wish IV.die +CA +3P +IND.1SG}\\
`I want to kill him' (lit. `I come to kill him')
\end{enumerate}
\normalsize

Under the prosodic-orthographic hypothesis, this is not a case of grammaticalisation from motion to volition, but of \textbf{translator interference}. The fact that \mapu{küpa} retains its motion semantics in independent use (\mapu{küpa-nu-l-i-iñ} `if we do not come', \hyperref[e03b]{E.03.b}) confirms its lexical vitality. The volitional reading is an artifact of translation, reinforced by the orthographic convention that presented \mapu{küpa} as a separate word—visually parallel to Spanish  \textit{querer} (`want') + infinitive.

\subsubsection{\label{sec.02.2.4} The Category-Neutral Nature of Concatenation}

The hypothesis that preverbal elements are ordinary verbs in V1 position gains crucial support from a broader observation about \mapu{Mapudüngun} verbal structure. The same verb-first concatenation pattern that places \mapu{pepi} before \mapu{tripa} in (\hyperref[e14a]{E.14.a}) also governs the incorporation of elements from other lexical categories. Consider the examples in (\hyperref[e14]{E.14}):

\paragraph{\example{14} \label{e14}}~
\vspace{-10pt}\begin{enumerate}[label=\alph*.]
\item \label{e14a} \mapuex{pepi \hspace{16pt} -tripa \hspace{12pt} -nu \hspace{8pt} -n} \hfill \citep{bcn-recurso-amparo2011}\\
      \gloss{TV.can IV.leave +NEG +PVN}\\
      `deprivation of liberty' (lit.: cannot leave) (verb-verb)
\item \label{e14b} \mapuex{küre \hspace{20pt} -nge \hspace{8pt} -n} \hfill (\citealp{smeets2008}: 194; 65)\\
     \gloss{NN.wife +IV.be +PVN}\\
     `to be married' (noun incorporation)
\item \label{e14c} \mapuex{weda \hspace{16pt} -künu \hspace{8pt} -y \hspace{8pt} -ø} \hfill (\citealp{smeets2008}: 409; 41)\\
     \gloss{AJ.bad +TV.put +IND +3}\\
     `they made things worse' (adjective incorporation)
\item \label{e14d} \mapuex{ina \hspace{38pt} -ye \hspace{16pt} -ngüma \hspace{4pt} -y \hspace{8pt} -ø \hspace{2pt} -iñ} \hfill (\citealp{smeets2008}: 209; 155)\\
     \gloss{AV.near +TV.carry +IV.cry +IND +1 +PL}\\
     `we cried with her' (adverb incorporation + verb concatenation)
\end{enumerate}
\normalsize

Three facts about these examples are immediately striking. First, the structural position of the modifying element is identical in each case: it directly precedes the verb it modifies, whether that modifier is a verb, noun, adjective, or adverb. Second, the category of the modifier varies freely, yet the construction remains the same. Third, elements that \cite{smeets2008} analyses as occupying fixed slots in a suffix template—such as \mapu{-nge} (slot 23, passive) and \mapu{-ye} (slot 35, oblique object)—reveal themselves under this analysis as ordinary verbs (\mapu{nge} `to be', \mapu{ye} `to carry') participating in the same concatenation system as \mapu{pepi} and \mapu{tripa}.

This category neutrality has profound implications. The preverbal elements that are the focus of this study are not a special auxiliary or modal class. They are simply verb roots that happen to appear with high frequency in the modifier position of a fully general concatenation system—a system that treats verbs, nouns, adjectives, and adverbs as equally eligible modifiers. The apparent hybridity that has puzzled previous analysts arises not from the grammar of these elements, but from the mismatch between \mapu{Mapudüngun}'s category-neutral concatenation and the category-bound categories of European linguistic description.

\subsubsection{\label{sec.02.2.5} The Spanish Contact Question}

Spanish influence is undeniable in the Contemporary period, but its primary effect has been \textbf{orthographic reinforcement} rather than syntactic restructuring. The analytic Spanish model—\textit{poder} (`can') + infinitive—validated the existing separated transcription convention, making it more persistent and, crucially, providing a template for the \textbf{syntactic reanalysis} evident in the Contemporary insertion data (\S\ref{sec.04.5}, \hyperref[e17]{E.17}–\hyperref[e19]{E.19}).

Without the prior orthographic convention, Spanish influence alone would not have produced the patterns we observe. A speaker who conceptualises \mapu{pepi} as part of a compound has no slot in which to insert \mapu{doy} `more' or \mapu{matu} `quickly'. But a speaker who has internalised the orthographic space as a real boundary—reinforced by decades of reading texts where \mapu{pepi} stands alone—may reinterpret that space as a syntactic position. The insertion evidence documents exactly this reanalysis in progress.

Thus, the interaction of two factors—orthographic fossilisation from the missionary period, plus contact with an analytic language that mirrored the resulting orthographic pattern—created the conditions for genuine syntactic change in the most recent period. The spaces first introduced by missionaries have, after four centuries, begun to become grammar.

\subsection{\label{sec.02.3} Diagnostic Predictions}

The prosodic-orthographic hypothesis makes specific, testable predictions about what should be found in the corpus. The following sections (\S\ref{sec.03}–\S\ref{sec.04}) test these predictions against a diachronic corpus spanning four centuries.

\subsubsection{\label{sec.02.3.1} Prediction 1: The V2/V3 Litmus Test}

If these elements are true lexical verbs participating in the category-neutral, verb-first concatenation system, they should occasionally appear as the \textbf{second or third element} in verbal compounds (e.g., \mapu{X-pepi-}, \mapu{X-kim-}, \mapu{X-küpa-}). Such V2 attestations would be incompatible with analyses that treat them as dedicated auxiliaries, prefixes, or particles—categories which, by definition, cannot occupy non-head positions in compounds.

The prediction is therefore straightforward: if even a handful of clear V2 examples can be documented across the corpus, the auxiliary/prefix/particle analyses are falsified. Section \ref{sec.04.2} presents the results of this test.

\subsubsection{\label{sec.02.3.2} Prediction 2: Orthographic Instability, Grammatical Stability}

If the prosodic-orthographic hypothesis is correct, the fusion rate (percentage of V1 tokens written as fused) should not correlate with grammatical behaviour. This follows from the fundamental independence of orthographic convention and grammatical structure: grammar changes slowly (over centuries, in consistent directions, affecting whole classes of elements), while orthography can shift rapidly (between editions, across authors, reflecting conventions rather than structural rules). Therefore, if the variation is orthographic rather than grammatical, we expect to see three specific patterns.

First, \textbf{decoupling}: stable or increasing V1 rates should appear alongside fluctuating fusion rates. If the underlying grammar is stable, the V1 rate should be stable or show gradual, directional change. But if orthographic convention is independent, the fusion rate can fluctuate without affecting how speakers actually use the element.

Second, \textbf{cross-element inconsistency}: roots with similar grammatical profiles should show no consistent orthographic trajectory. If fusion and separation reflected grammatical status, elements with the same syntactic behaviour would be expected to show the same orthographic pattern. But if orthography is a matter of convention, different elements can acquire different writing conventions regardless of their grammar.

Third, \textbf{abrupt shifts}: dramatic orthographic shifts between adjacent periods should occur without corresponding changes in grammatical position. Grammatical change is slow, requiring generations for syntactic structures to shift; orthographic change can be abrupt.

These patterns are examined in \S\ref{sec.04.4}, where we present the fusion rate analysis.

\subsubsection{\label{sec.02.3.3} Prediction 3: Phonological Coherence}

Truly fused forms should exhibit a higher incidence of \textbf{morphophonological processes} (vowel harmony, assimilation, elision) across the morpheme boundary than artificially separated forms. If the spaces in separated forms represent genuine prosodic boundaries, they should inhibit such processes; if they are orthographic only, we expect no systematic difference.

This prediction remains to be tested systematically in future research, though impressionistic evidence from the corpus—such as the preservation of root integrity in both fused and separated forms—suggests that phonological processes operate independently of orthography.

\begin{quote}
\small
\textbf{How This Could Be Tested:} A systematic phonological analysis of the boundary between the preverbal root and the following verb would test this prediction. Three types of evidence could be examined:

\textbf{First, vowel harmony:} \mapu{Mapudüngun} exhibits vowel harmony in certain compound contexts (e.g., \mapu{küpa} + \mapu{yiñ} → \mapu{küpa-yi-ñ} where the root vowel influences the suffix). If fusion is genuine, we would expect harmony to apply across the boundary in fused forms but not in separated forms. If the space is orthographic only, harmony should apply regardless of spelling.

\textbf{Second, consonant assimilation:} In true compounds, root-final consonants may assimilate to the following consonant (e.g., \mapu{pepi} + \mapu{tripa} → \mapu{pepi-tripa} with no assimilation because \mapu{/p/} and \mapu{/t/} are both stops; but cases with nasal-stop sequences could show assimilation). Separated forms should inhibit such assimilation if the space represents a genuine prosodic boundary.

\textbf{Third, elision:} In rapid speech, vowels may be elided at morpheme boundaries in true compounds. If the space is orthographic only, elision should occur regardless of spelling; if the space represents a prosodic boundary, elision should be blocked.

A controlled study would compare minimal pairs: the same root combinations appearing in both fused and separated orthography in the same text or by the same author. Acoustic analysis could measure pause duration, vowel formants, and consonant closure durations to determine whether the orthographic space corresponds to a measurable phonetic boundary. Such a study remains for future research.
\end{quote}
\normalsize

\subsubsection{\label{sec.02.3.4} Prediction 4: The Insertion Diagnostic}

If the separation has been reanalysed as syntactically real in the Contemporary period, we expect to find \textbf{material intervening} between the preverbal element and the main verb in recent texts—adverbs, quantifiers, even other verbs. Such insertion should be absent or extremely rare in Early texts, appearing only after the orthographic convention has had time to influence speaker intuitions.

Section \ref{sec.04.5} documents this pattern, presenting the insertion data in \hyperref[tab04]{Table 4}.

\subsubsection{\label{sec.02.3.5} Prediction 5: Intra-Author Variation}

If the variation is orthographic rather than grammatical, the \textbf{same author} in the \textbf{same text} should sometimes use fused forms and sometimes separated forms for the same element, with no consistent functional distinction. Such variation would be inexplicable under a syntactic analysis but expected under an orthographic one.

This prediction is examined in \S\ref{sec.04.5}, where we present evidence from 1990s CONADI documents, where \mapu{pepi} appears both fused and separated in the same passages with no discernible difference in meaning or grammatical function.

\section{\label{sec.03} Methodology}

This study employs a tripartite corpus-driven framework to test the predictions of the prosodic-orthographic hypothesis formulated in \S\ref{sec.02}. The methodology combines (i) diachronic corpus design, (ii) systematic tagging and quantification, and (iii) targeted diagnostic analyses of specific grammatical phenomena.

\subsection{\label{sec.03.1} Corpus Design and Periodisation}

To track changes in orthographic practice and grammatical structure over time, the corpus is divided into three periods corresponding to major phases in the documentation of \mapu{Mapudüngun}:

\begin{enumerate}
\item \textbf{Early Period (1606–1846)}: From the first missionary grammar (de Valdivia 1606) through the influential work of \cite{febres1765} and its subsequent editions. This period represents the initial fixation of orthographic conventions, with texts produced by European missionaries with native speaker input.

\item \textbf{Transition Period (1916–1930)}: Centred on de Augusta's dictionary (1916) and the autobiography of \mapu{Paskwal Koña} \citep{mosbach1930}. This period includes the first extended texts produced by native speakers working with European ethnographers, offering a bridge between missionary transcription and contemporary usage.

\item \textbf{Contemporary Period (1992–present)}: From \cite{salas1992a} through the present, encompassing academic grammars (\citealp{smeets2008}; \citealp{zuniga2006}), native-authored poetry and narrative (\citealp{wirimilla2005b}; \citealp{marileo2006}), and institutional translations (\citealp{conadi1994}--\citeyear{conadi1996b}; \citealp{convenio169oit2006}; \citealp{bcn-credito-universitario2006}--\citeyear{bcn2023}).
\end{enumerate}

The corpus includes both dictionary entries (which provide isolated examples) and connected texts (which preserve naturalistic usage). The dictionary sources are drawn from the CORLEXIM database \citep{chandia2014} and the author's own compilation of published materials.

\textbf{Corpus size.} The complete corpus comprises approximately 1.26 million running words across all periods. The Early period contributes roughly 130,000 words, the Transition period approximately 360,000 words, and the Contemporary period approximately 770,000 words. These totals include both dictionary entries and connected texts; dictionary entries were counted at the level of the full entry (including definitions and examples) to reflect their contribution to the total textual material analysed. The variation in period sizes reflects the historical growth of documentation: the missionary period produced relatively few texts, while the Contemporary period has seen a substantial expansion in published materials, including institutional translations, educational resources, and native-authored literature.

All sources used in this corpus are published materials or accessed through the CORLEXIM database \citep{chandia2014}, which provides scholarly access to historical and contemporary \mapu{Mapudüngun} dictionaries. No fieldwork with human subjects was conducted for this study.

\subsection{\label{sec.03.2} Data Extraction and Tagging}

For each of the six target roots—\mapu{kalli}, \mapu{kim}, \mapu{küpa}, \mapu{pepi}, \mapu{shinge}, and \mapu{wüño}—all occurrences were extracted from the corpus and organised in periodised files. Each token was tagged for the following variables:

\begin{itemize}
\item \textbf{Source}: text identifier, year, author, genre (dictionary/grammar/narrative/translation)
\item \textbf{Variant}: orthographic form attested (e.g., \mapu{wüño}, \mapu{wiño}, \mapu{wüno})
\item \textbf{Context}: the surrounding text, sufficient to determine grammatical environment
\item \textbf{Position}: classified as:
    \begin{itemize}
    \item \textbf{V1}: root appears as first element in a verbal compound
    \item \textbf{V2}: root appears as second element in a verbal compound (crucial for testing Prediction 1, \S\ref{sec.02.3.1})
    \item \textbf{Isolated}: root does not appear in a verbal compound (as independent predicate, noun, etc.)
    \end{itemize}
\item \textbf{Fusion status} (for V1 tokens): whether written as a single orthographic word (fused, e.g., \mapu{pepi-weyel-}) or with a space (separated, e.g., \mapu{pepi weyel})
\item \textbf{Insertion}: presence of any material intervening between the target root and the following verb
\end{itemize}

Tagging was performed manually by the author, with ambiguous cases resolved through consultation of source context.

A note on the treatment of \mapu{wüño} is necessary. In \mapu{Mapudüngun}, \mapu{wüño} also refers to the wooden stick used in the game of \mapu{palin} (a sport similar to field hockey). The connection between the two meanings is transparent: the stick is used to `return' the ball. To avoid conflating the nominal and verbal uses, all instances of \mapu{wüño} referring to the stick were excluded from the analysis. Only tokens where \mapu{wüño} functions as a verb or verbal modifier were retained.

The complete tagged dataset is available in the author's repository under the title \textit{MPRDataset: Mapudüngun Preverbal Roots Dataset} \citep{chandia2026}. Access is provided under a Creative Commons CC BY-NC-SA 4.0 licence at \url{https://www.chandia.net/dungupeyum/repositorio}.

\subsection{\label{sec.03.3} Quantitative Analysis}

The tagged data were analysed using simple descriptive statistics to address the following research questions:

\begin{enumerate}
\item What is the frequency of V1, V2, and isolated occurrences for each root in each period?
\item How does the fusion rate (percentage of V1 tokens written as fused) change across periods for each root?
\item What is the rate of V2 attestations for each root, and what does this imply about their lexical status?
\item When does insertion first appear, and which roots permit it?
\end{enumerate}

Results are presented in \S\ref{sec.04}, organised by diagnostic prediction.

\subsection{\label{sec.03.4} Diagnostic Tests}

Beyond simple quantification, four targeted analyses were conducted to test specific predictions of the prosodic-orthographic hypothesis.

\subsubsection{\label{sec.03.4.1} The V2/V3 Test}

All V2 attestations were manually examined to confirm that the target root indeed functions as the second element in a verbal compound (e.g., \mapu{X-kim-}), not as an independent verb in a separate clause. Each V2 token was documented with full context to support qualitative discussion in \S\ref{sec.04.2}.

\subsubsection{\label{sec.03.4.2} The Insertion Test}

All tokens were searched for intervening material between the target root and the following verb. Following the diagnostic established in \S\ref{sec.02.3.4}, such insertion is expected to be absent or rare in Early texts and increasingly frequent in Contemporary texts if orthographic conventions have been reanalysed as syntactic boundaries.

A crucial methodological distinction must be drawn between \textbf{true insertion} and \textbf{separated incorporation}. True insertion occurs when an element appears between V1 and V2 that cannot be analysed as an incorporated modifier of the following verb—typically adverbs like \mapu{doy} `more' or demonstratives like \mapu{fey} `that'. Separated incorporation, by contrast, occurs when an element that can appear fused to the following verb in other attestations (e.g., \mapu{küme} `good', \mapu{pichi} `little') is written with a space due to orthographic convention rather than syntactic independence. The diagnostic for true insertion is the absence of fused forms in the same period: if an element never appears incorporated into the following verb, its appearance between V1 and V2 constitutes genuine syntactic insertion.

This distinction is essential for testing the prosodic-orthographic hypothesis. Separated incorporation is predicted from the earliest periods—it is simply the orthographic representation of the same compounding structure that also appears fused. True insertion, by contrast, is predicted to emerge only in the Contemporary period, as the orthographic space is reanalysed as a genuine syntactic boundary.

\subsubsection{\label{sec.03.4.3} Fusion Rate Analysis}

For V1 tokens, fusion rates were calculated per period and per root. According to Prediction 2 (\S\ref{sec.02.3.2}), these rates should not show a linear historical trend; instead, they should correlate with text genre and translator identity. Rates were therefore also calculated separately for different text types within each period.

\subsubsection{\label{sec.03.4.4} Intra-Author Variation}

Where the same author appears multiple times in the corpus (e.g., CONADI documents from the 1990s), fused and separated forms of the same root were compared within single texts. Following Prediction 5 (\S\ref{sec.02.3.5}), such variation, if present, would support an orthographic rather than grammatical account.

\subsection{\label{sec.03.5} Limitations}

Several limitations should be acknowledged. First, the corpus is necessarily restricted to written sources; we have no access to spoken usage before the late twentieth century. Second, the Early period texts are exclusively the work of European missionaries, whose transcriptions may reflect perceptual biases discussed in \S\ref{sec.02.1}. Third, the quantity of data varies considerably across periods and roots (see \hyperref[tab02]{Table 2}), with some cells containing too few tokens for robust statistical comparison. Where token counts are low (N < 10), results are interpreted with caution and noted as such in the discussion.

Despite these limitations, the corpus provides the most comprehensive diachronic dataset available for \mapu{Mapudüngun}, and the consistency of the patterns observed across multiple roots and periods (see \S\ref{sec.04}) suggests that the findings are robust.

\section{\label{sec.04} Results}

This section presents the quantitative and qualitative findings of the corpus analysis, organised according to the diagnostic predictions formulated in \S\ref{sec.02.3}. Section \ref{sec.04.1} provides a quantitative overview of the positional distribution of each root across the three historical periods. Sections \ref{sec.04.2}–\ref{sec.04.5} then examine specific patterns in detail: the crucial V2 attestations that confirm verbal status (\S\ref{sec.04.2}), the distinctive behaviour of \mapu{kalli} (\S\ref{sec.04.3}), the fusion rate analysis that reveals orthographic variation (\S\ref{sec.04.4}), and the insertion evidence that demonstrates the productivity of the concatenation system (\S\ref{sec.04.5}). Together, these findings provide empirical support for the prosodic-orthographic hypothesis.

\subsection{\label{sec.04.1} Quantitative Overview}

Two complementary perspectives on the data are presented in \hyperref[tab02]{Table 2} and \hyperref[tab03]{Table 3}.

\hyperref[tab02]{Table 2} tracks \textbf{orthographic representation}—whether writers chose to fuse the preverbal element with the following verb (e.g., \mapu{pepi-weyel-}) or write it separately (e.g., \mapu{pepi weyel}). \hyperref[tab03]{Table 3} tracks \textbf{grammatical position}—whether the element functions as the first (V1) or second (V2) element in a verbal compound, or appears outside a verbal compound entirely (Isolated), either as an independent inflected verb or in other constructions.

\subsubsection{\label{sec.04.1.1} Orthographic Patterns (\hyperref[tab02]{Table 2})}

\hyperref[tab02]{Table 2} reveals striking variation in writing conventions across roots and periods. \mapu{Kalli} is overwhelmingly written separately in all periods (87–98\% separated), presenting an orthographic profile consistent with a particle. \mapu{Kim}, by contrast, is almost categorically fused in Early and Transition texts (99\%), with only a slight increase in separated forms in the Contemporary period (3.6\%). \mapu{Küpa} shows a U-shaped pattern: fusion at 78\% in Early, rising to 94\% in Transition, then dropping back to 77\% in Contemporary.

The most dramatic pattern is exhibited by \mapu{pepi}. Its fusion rate remains stable through Early and Transition periods ($\sim$65\%), then \textbf{collapses to 37.5\%} in Contemporary texts—a shift of nearly 30 percentage points. \mapu{Wüño} shows a different trajectory: near-even split in Transition (52\% fused), with fusion recovering to 77\% in Contemporary. \mapu{Shinge} is too sparsely attested for reliable analysis.

\begin{longtable}{@{}p{1.8cm}p{1.6cm}ccccc>{\raggedright\arraybackslash}p{3.8cm}@{}}
\caption{Orthographic Representation of Preverbal Roots Across Historical Periods}
\label{tab02}\\
\toprule
\textbf{Root} & \textbf{Period} & \textbf{Fused} & \textbf{Sep.} & \textbf{Fused} & \textbf{Sep.} & \textbf{Total} & \textbf{Key Interpretation}\\
& & {\footnotesize N} & {\footnotesize N} & {\footnotesize(\%)} & {\footnotesize(\%)} & \textbf{N} & \\
\midrule
\endfirsthead
\multicolumn{8}{c}{\tablename\ \thetable\ -- \textit{Continued}}\\
\toprule
\textbf{Root} & \textbf{Period} & \textbf{Fused} & \textbf{Sep.} & \textbf{Fused} & \textbf{Sep.} & \textbf{Total} & \textbf{Key Interpretation}\\
& & {\footnotesize N} & {\footnotesize N} & {\footnotesize(\%)} & {\footnotesize(\%)} & \textbf{N} & \\
\midrule
\endhead
\bottomrule
\endfoot
\bottomrule
\endlastfoot
\mapu{\textbf{kalli}} & Early & 4 & 90 & 4.3 & 95.7 & 94 & Overwhelmingly written separated. \\
\scriptsize(`let, allow') & Transit. & 1 & 49 & 2.0 & 98.0 & 50 & Orthographic convention stable. \\
& Contemp. & 2 & 14 & 12.5 & 87.5 & 16 & Slight fusion increase. \\
\midrule
\mapu{\textbf{kim}} & Early & 380 & 3 & 99.2 & 0.8 & 383 & \textbf{Near-total fusion}. \\
\scriptsize(`know how') & Transit. & 1119 & 3 & 99.7 & 0.3 & 1122 & Orthographic convention stable. \\
& Contemp. & 1404 & 52 & 96.4 & 3.6 & 1456 & Slight rise in separation. \\
\midrule
\mapu{\textbf{küpa}} & Early & 162 & 45 & 78.3 & 21.7 & 207 & Majority fused. \\
\scriptsize(`want/wish') & Transit. & 588 & 36 & 94.2 & 5.8 & 624 & \textbf{Shift to fusion}. \\
& Contemp. & 317 & 95 & 76.9 & 23.1 & 412 & Returns to Early pattern. \\
\midrule
\mapu{\textbf{pepi}} & Early & 117 & 64 & 64.6 & 35.4 & 181 & Mixed pattern. \\
\scriptsize(`can') & Transit. & 192 & 99 & 66.0 & 34.0 & 291 & Stable mixed pattern. \\
& Contemp. & 162 & 270 & 37.5 & 62.5 & 432 & \textbf{Separation dominates}. \\
\midrule
\mapu{\textbf{wüño}} & Early & 6 & 3 & 66.7 & 33.3 & 9 & Low N. \\
\scriptsize(`return/re-') & Transit. & 156 & 142 & 52.3 & 47.7 & 298 & \textbf{Near-even split}. \\
& Contemp. & 157 & 48 & 76.6 & 23.4 & 205 & Returns to fusion. \\
\midrule
\mapu{\textbf{shinge}} & Early & 0 & 0 & — & — & 0 & No data. \\
\scriptsize(`move along') & Transit. & 46 & 0 & 100.0 & 0.0 & 46 & Exclusively fused. \\
& Contemp. & 1 & 0 & 100.0 & 0.0 & 1 & Too sparse. \\
\bottomrule
\end{longtable}
\begin{center}
\begin{minipage}{12cm}
\footnotesize Note: Fused = written as single orthographic word; Sep. = Separated (space between root and following element). `—' indicates no data. \\
\end{minipage}
\end{center}

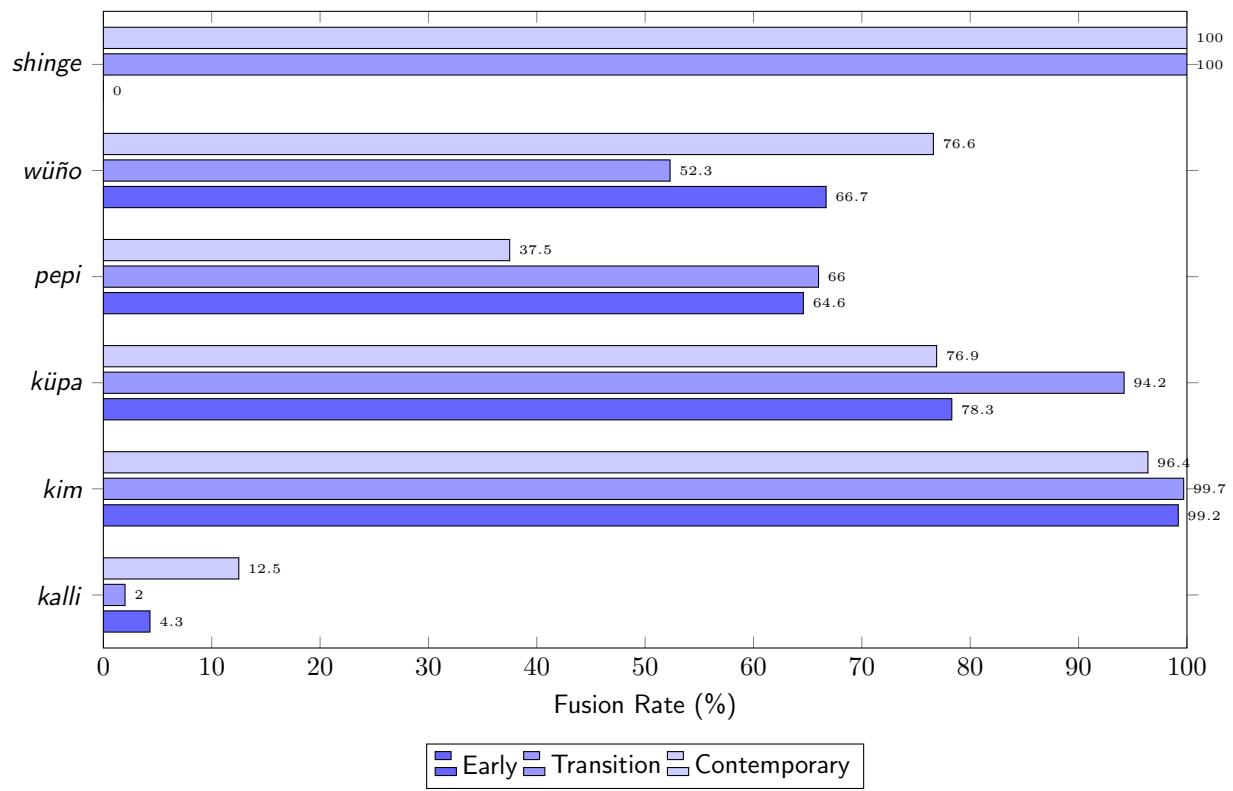
\begin{figure}[H]
\centering
\begin{tikzpicture}
\begin{axis}[
    xbar,
    xlabel={Fusion Rate (\%)},
    ytick={1,2,3,4,5,6},
    yticklabels={
        \mapu{kalli},
        \mapu{kim},
        \mapu{küpa},
        \mapu{pepi},
        \mapu{wüño},
        \mapu{shinge}
    },
    xmin=0, xmax=100,
    legend style={at={(0.5,-0.15)}, anchor=north, legend columns=3},
    bar width=0.2,
    width=\linewidth,
    height=10cm,
    nodes near coords,
    nodes near coords align={horizontal},
    every node near coord/.append style={font=\tiny}
]
\addplot [fill=blue!60] coordinates {(4.3,1) (99.2,2) (78.3,3) (64.6,4) (66.7,5) (0,6)};
\addplot [fill=blue!40] coordinates {(2.0,1) (99.7,2) (94.2,3) (66.0,4) (52.3,5) (100.0,6)};
\addplot [fill=blue!20] coordinates {(12.5,1) (96.4,2) (76.9,3) (37.5,4) (76.6,5) (100.0,6)};
\legend{Early, Transition, Contemporary}
\end{axis}
\end{tikzpicture}
\caption{Fusion Rates by Period and Root (\% of V1 tokens written as fused)}
\label{fig:01}
\end{figure}

\hyperref[fig:01]{Figure 1} visualises the fusion rate data from \hyperref[tab02]{Table 2}, highlighting the dramatic patterns across periods and roots—particularly the collapse in \mapu{pepi}'s fusion rate in the Contemporary period and the near-total fusion of \mapu{kim} across all periods.\\

\begin{quote}
\small
\textbf{Note on \mapu{shinge}:} The 100\% fusion rate for \mapu{shinge} in the Transition and Contemporary periods should be interpreted with caution. This figure reflects only 47 attestations in the Transition period and a single attestation in the Contemporary period. Unlike \mapu{kim}, which maintains near-total fusion across thousands of tokens, \mapu{shinge}'s apparent categorical fusion is an artifact of sparse data rather than a robust grammatical pattern. We return to the implications of this sparsity in \S\ref{sec.04.3}.\\
\end{quote}
\normalsize

\subsubsection{\label{sec.04.1.2} Grammatical Patterns (\hyperref[tab03]{Table 3})}

Beneath this orthographic variation, \hyperref[tab03]{Table 3} reveals remarkable grammatical stability. All six elements function as V1 modifiers, appear in the crucial V2 position (confirming their status as full lexical verbs), and occur as isolated inflected verbs.

\mapu{Kalli} is primarily a V1 modifier (68–87\%) across all periods—a grammatical fact obscured by its consistently separated orthography. \mapu{Kim} and \mapu{küpa} show strong preferences for isolation (55–83\%), behaving as independent verbs that happen also to occur in V1 position. Their V2 attestations—36 for \mapu{kim}, 42 for \mapu{küpa}—are conclusive proof that they are ordinary verbs, not a special grammatical class.

\mapu{Pepi} provides the clearest evidence for the prosodic-orthographic hypothesis. Its V1 rate \textbf{increases} steadily across periods, from 41\% in Early to \textbf{82\% in Contemporary}, even as its fusion rate collapses (as shown in \hyperref[fig:01]{Figure 1}). The grammar is moving in the opposite direction from orthography: \mapu{pepi} is used as a V1 modifier more than ever, but written separately more than ever.

\mapu{Wüño} presents a more complex picture: V1 rate peaks in Transition (74\%) then drops to 42\% in Contemporary, with one crucial V2 attestation in the Contemporary period confirming its verbal status.

\begin{longtable}{@{}p{1.8cm}p{1.6cm}ccccl>{\raggedright\arraybackslash}p{3.8cm}@{}}
\caption{Grammatical Position of Preverbal Roots Across Historical Periods}
\label{tab03}\\
\toprule
\textbf{Root} & \textbf{Period} & \textbf{V1} & \textbf{V2} & \textbf{Iso.} & \textbf{Total} & \textbf{Key Interpretation}\\
& & {\footnotesize(\%)} & {\footnotesize(\%)} & {\footnotesize(\%)} & \textbf{N} & \\
\midrule
\endfirsthead
\multicolumn{7}{c}{\tablename\ \thetable\ -- \textit{Continued}}\\
\toprule
\textbf{Root} & \textbf{Period} & \textbf{V1} & \textbf{V2} & \textbf{Iso.} & \textbf{Total} & \textbf{Key Interpretation}\\
& & {\footnotesize(\%)} & {\footnotesize(\%)} & {\footnotesize(\%)} & \textbf{N} & \\
\midrule
\endhead
\bottomrule
\endfoot
\bottomrule
\endlastfoot
\mapu{\textbf{kalli}} & Early & 83.0 & 0.0 & 17.0 & 94 & \textbf{Primarily V1} despite orthographic separation. \\
\scriptsize(`let, allow') & Transit. & 68.6 & 3.9 & 27.5 & 51 & V1 dominant; rare V2. \\
& Contemp. & 87.5 & 0.0 & 12.5 & 16 & V1 increases. \\
\midrule
\mapu{\textbf{kim}} & Early & 29.5 & 3.2 & 67.4 & 380 & \textbf{More isolated than V1}—full verb. \\
\scriptsize(`know how') & Transit. & 39.0 & 1.1 & 60.0 & 1109 & V1 increases slightly. \\
& Contemp. & 44.1 & 0.8 & 55.1 & 1450 & V1 rising; still mainly isolated. \\
\midrule
\mapu{\textbf{küpa}} & Early & 22.3 & 2.9 & 74.8 & 206 & \textbf{Strongly prefers isolation}. \\
\scriptsize(`want/wish') & Transit. & 11.0 & 5.1 & 83.8 & 625 & V1 drops; V2 peaks. \\
& Contemp. & 32.8 & 1.0 & 66.3 & 415 & V1 recovers. \\
\midrule
\mapu{\textbf{pepi}} & Early & 41.4 & 1.1 & 57.5 & 181 & Mixed pattern. \\
\scriptsize(`can') & Transit. & 47.4 & 0.0 & 52.6 & 291 & Stable mixed pattern. \\
& Contemp. & \textbf{82.3} & 0.0 & 17.7 & 429 & \textbf{V1 doubles} as fusion drops (\hyperref[fig:01]{Figure 1}). \\
\midrule
\mapu{\textbf{wüño}} & Early & 33.3 & 0.0 & 66.7 & 9 & Low N. \\
\scriptsize(`return/re-') & Transit. & 73.6 & 0.0 & 26.4 & 295 & \textbf{Shift to V1}. \\
& Contemp. & 41.7 & 0.5 & 57.8 & 204 & Returns to isolation; crucial V2. \\
\midrule
\mapu{\textbf{shinge}} & Early & — & — & — & 0 & No data. \\
\scriptsize(`move along') & Transit. & 84.8 & 0.0 & 15.2 & 46 & Exclusively V1. \\
& Contemp. & 100.0 & 0.0 & 0.0 & 1 & Too sparse. \\
\bottomrule
\end{longtable}
\begin{center}
\begin{minipage}{12cm}
\footnotesize Note: V1 = first element in verb complex; V2 = second element; Iso. = isolated—outside verb complex or inflected verb with no root incorporation. V2 attestations: \textit{kalli} (2), \textit{kim} (36), \textit{küpa} (42), \textit{pepi} (2), \textit{wüño} (1)—confirming full lexical verb status. `—' indicates no data.\\
\end{minipage}
\end{center}

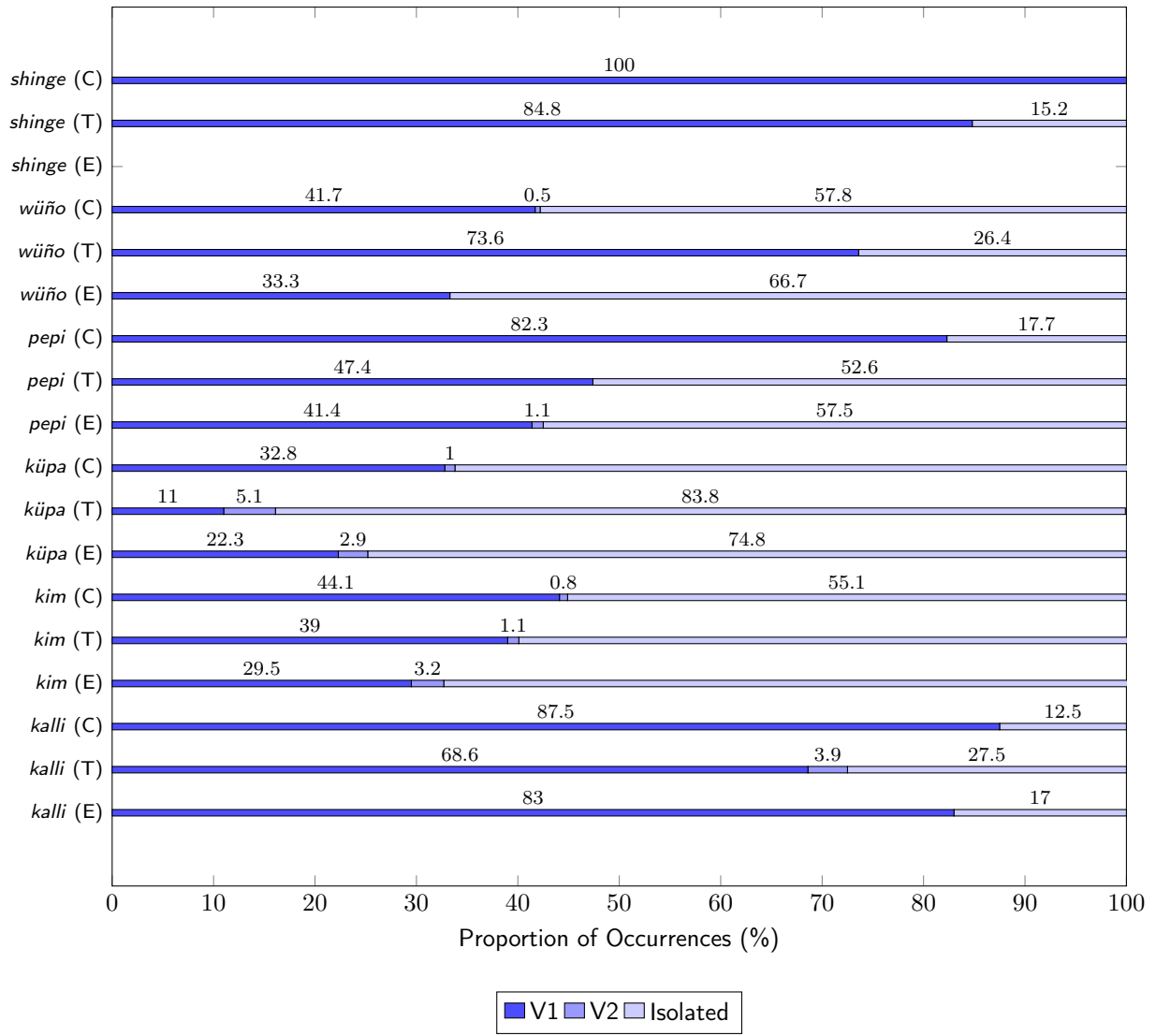
\begin{figure}[H]
\centering
\begin{tikzpicture}
\begin{axis}[
    xbar stacked,
    xlabel={Proportion of Occurrences (\%)},
    ytick={1,2,3,4,5,6,7,8,9,10,11,12,13,14,15,16,17,18},
    yticklabels={
        \mapu{kalli} (E),
        \mapu{kalli} (T),
        \mapu{kalli} (C),
        \mapu{kim} (E),
        \mapu{kim} (T),
        \mapu{kim} (C),
        \mapu{küpa} (E),
        \mapu{küpa} (T),
        \mapu{küpa} (C),
        \mapu{pepi} (E),
        \mapu{pepi} (T),
        \mapu{pepi} (C),
        \mapu{wüño} (E),
        \mapu{wüño} (T),
        \mapu{wüño} (C),
        \mapu{shinge} (E),
        \mapu{shinge} (T),
        \mapu{shinge} (C)
    },
    xmin=0, xmax=100,
    legend style={at={(0.5,-0.12)}, anchor=north, legend columns=3},
    bar width=0.15,
    width=\linewidth,
    height=14cm,
    nodes near coords,
    nodes near coords align={vertical},
    every node near coord/.append style={font=\footnotesize},
    yticklabel style={font=\footnotesize}
]
\addplot [fill=blue!70] coordinates {
    (83.0,1) (68.6,2) (87.5,3)
    (29.5,4) (39.0,5) (44.1,6)
    (22.3,7) (11.0,8) (32.8,9)
    (41.4,10) (47.4,11) (82.3,12)
    (33.3,13) (73.6,14) (41.7,15)
    (0,16) (84.8,17) (100.0,18)
};
\addplot [fill=blue!40] coordinates {
    (0.0,1) (3.9,2) (0.0,3)
    (3.2,4) (1.1,5) (0.8,6)
    (2.9,7) (5.1,8) (1.0,9)
    (1.1,10) (0.0,11) (0.0,12)
    (0.0,13) (0.0,14) (0.5,15)
    (0,16) (0.0,17) (0.0,18)
};
\addplot [fill=blue!20] coordinates {
    (17.0,1) (27.5,2) (12.5,3)
    (67.4,4) (60.0,5) (55.1,6)
    (74.8,7) (83.8,8) (66.3,9)
    (57.5,10) (52.6,11) (17.7,12)
    (66.7,13) (26.4,14) (57.8,15)
    (0,16) (15.2,17) (0.0,18)
};
\legend{V1, V2, Isolated}
\end{axis}
\end{tikzpicture}
\caption{Grammatical Position by Period and Root (\% of occurrences); E: Early, T: Transition, C: Contemporary}
\label{fig:02}
\end{figure}

\hyperref[fig:02]{Figure 2} visualises the grammatical position data from Table 3, highlighting the dramatic increase in \mapu{pepi}'s V1 rate in the Contemporary period and the stable V1 dominance of \mapu{kalli} across all periods.\\

\begin{quote}
\small
\textbf{Note on \mapu{shinge}:} The 100\% V1 rate for \mapu{shinge} in the Contemporary period reflects only a single attestation. The Transition period figure (84.8\%) is based on 46 attestations, all from de Augusta's dictionary. The sparsity of data for \mapu{shinge} (see \hyperref[tab03]{Table 3}) means that its profile should not be interpreted as robust evidence for grammatical behaviour; we return to this point in \S\ref{sec.04.3}.\\
\end{quote}
\normalsize

The contrast between the two tables—and their corresponding figures—is immediately revealing. Consider \mapu{pepi} in the Contemporary period: \hyperref[tab02]{Table 2} and \hyperref[fig:01]{Figure 1} show it is fused only 37.5\% of the time, yet \hyperref[tab03]{Table 3} and \hyperref[fig:02]{Figure 2} show it functions as V1 in 82.3\% of its occurrences. This disjuncture—high grammatical V1 usage alongside low orthographic fusion—is exactly what the prosodic-orthographic hypothesis predicts. Writers have internalised the missionary convention of spacing, but the underlying grammar remains verb-first concatenation.

\subsection{\label{sec.04.2} The V2/V3 Evidence}

The most direct test of whether these elements are full lexical verbs—as opposed to dedicated auxiliaries, prefixes, or particles—is their ability to appear as the \textbf{second element} in a verbal compound. Under radical concatenation, any verb root should be able to occupy V2 position; under the alternative analyses, this should be impossible.

\hyperref[tab03]{Table 3} provides clear evidence. Across the corpus, V2 attestations are documented for:

\begin{itemize}
\item \mapu{kalli}: 2 (Transition period)
\item \mapu{kim}: 36 (all periods)
\item \mapu{küpa}: 42 (all periods)
\item \mapu{pepi}: 2 (Early period)
\item \mapu{wüño}: 1 (Contemporary period)
\end{itemize}

Example (\hyperref[e15]{E.15}) illustrates the pattern:

\paragraph{\example{15} \label{e15}}~
\vspace{-10pt}\begin{enumerate}[label=\alph*.]
\item \label{e15a} \mapuex{af \hspace{26pt} -kalli \hspace{10pt} -a \hspace{10pt} -lu} \hfill (\citealp{deaugusta1910}: 339)\\
\gloss{NN.end +IV.let +FUT +SVN}\\
`solo así se acabarán' (lit. `will end letting')
\item \label{e15b} \mapuex{kecha \hspace{20pt} -küpa \hspace{10pt} -l \hspace{10pt} -tu \hspace{10pt} -n} \hfill (de Augusta, 1916 in \citealp{chandia2014})\\
\gloss{TV.urge +IV.come +CA +APL +PVN}\\
`urge [animals] on coming'
\end{enumerate}
\normalsize

These V2 attestations are not isolated anomalies. They span all three periods and involve multiple roots, demonstrating that these elements retain the ability to function as ordinary verbs in any position within the verbal compound. This finding is incompatible with analyses that treat them as dedicated auxiliaries, prefixes, or particles.

\citeauthor{deaugusta1910} (\citeyear{deaugusta1910}: 339) explicitly notes the equivalence: \mapu{afkallialu} = \mapu{kalli afalu} o \mapu{kalli afpe}. This meta-linguistic comment from a \mapu{Mapudüngun} speaker-linguist indicates that the fused and separated forms were perceived as variants of the same construction—exactly what the prosodic-orthographic hypothesis predicts.

\subsection{\label{sec.04.3} The \mapu{kalli} Anomaly}

\hyperref[tab02]{Table 2} shows that \mapu{kalli} is overwhelmingly written separately across all periods (87–98\% separated). Its orthographic profile is that of a particle. Yet \hyperref[tab03]{Table 3} reveals that \mapu{kalli} is \textbf{grammatically a V1 modifier} in 68–87\% of its occurrences. The disjuncture could not be starker: writers treat \mapu{kalli} as a separate word, but grammatically it functions as the first element in a verbal compound.

This pattern is exactly what the prosodic-orthographic hypothesis predicts. The missionary transcription convention introduced a space before the following verb, and that convention has persisted. But the underlying grammar—\mapu{kalli} as V1 in a verb-first concatenation—remains unchanged.

The two V2 attestations of \mapu{kalli} (Transition period) confirm that it is also a full lexical verb capable of occupying any position in a compound. \mapu{Kalli} is not a grammatical exception; it is an orthographic one.

\subsection{\label{sec.04.4} Fusion Rate Analysis}

If the prosodic-orthographic hypothesis is supported, fusion rates (\hyperref[tab02]{Table 2}) should not correlate with grammatical V1 rates (\hyperref[tab03]{Table 3}). Instead, orthography should fluctuate independently of grammar, reflecting text genre, translator identity, and shifting conventions rather than grammatical change.

\mapu{Pepi} provides the clearest test. Its V1 rate in \hyperref[tab03]{Table 3} rises steadily: 41\% (Early) → 47\% (Transition) → 82\% (Contemporary). Yet its fusion rate in \hyperref[tab02]{Table 2} and \hyperref[fig:01]{Figure 1} moves in the opposite direction: 65\% → 66\% → 37\%. The grammar is stabilising \mapu{pepi} as a V1 modifier, while orthography increasingly represents it as separate. This decoupling is inexplicable under a grammatical change account, but expected under orthographic fossilisation.

\mapu{Wüño} shows a different pattern: V1 rate peaks in Transition (74\%) then drops (42\%), while fusion rate dips in Transition (52\%) then recovers (77\%). Again, grammar and orthography move independently.

\mapu{Kim} and \mapu{küpa} show greater alignment, but this is because their orthographic conventions were fixed early and remained stable. The crucial point is that where orthography shifts, grammar does not follow.

The fusion patterns in \hyperref[tab02]{Table 2} and \hyperref[fig:01]{Figure 1} correlate closely with the insertion patterns in \hyperref[tab04]{Table 4}: the roots with the most dramatic fusion collapse (\mapu{pepi}) are precisely those that innovate true insertion in the Contemporary period.

\subsection{\label{sec.04.5} The Insertion Evidence: Syntactic Reanalysis in Progress}

A crucial diagnostic for the grammatical status of preverbal elements is whether other material can intervene between them and the main verb. True auxiliaries, prefixes, and particles should all resist such insertion; V1 elements in compounds, by contrast, participate in a verb-first concatenation system where modifiers naturally follow the verb they modify.

A methodological distinction must be drawn, however, between \textbf{true insertion} and \textbf{separated incorporation}. True insertion occurs when an element appears between V1 and V2 that cannot be analysed as an incorporated modifier of the following verb—typically adverbs like \mapu{doy} `more' or demonstratives like \mapu{fey} `that'. Separated incorporation, by contrast, occurs when an element that can appear fused to the following verb in other attestations (e.g., \mapu{küme} `good', \mapu{pichi} `little') is written with a space due to orthographic convention rather than syntactic independence. The diagnostic for true insertion is the absence of fused forms in the same period: if an element never appears incorporated into the following verb, its appearance between V1 and V2 constitutes genuine syntactic insertion.

\hyperref[tab04]{Table 4} presents the quantified distribution of insertion across periods, applying this distinction.

\begin{longtable}{@{}p{1.6cm}p{1.4cm}cccl>{\raggedright\arraybackslash}p{3.2cm}@{}}
\caption{Insertion Attestations by Period and Type}
\label{tab04}\\
\toprule
\textbf{Root} & \textbf{Period} & \textbf{True Ins.} & \textbf{Sep. Inc.} & \textbf{Total} & \textbf{Key Interpretation}\\
& & {\footnotesize N} & {\footnotesize N} & \textbf{N} & \\
\midrule
\endfirsthead
\multicolumn{6}{c}{\tablename\ \thetable\ -- \textit{Continued}}\\
\toprule
\textbf{Root} & \textbf{Period} & \textbf{True Ins.} & \textbf{Sep. Inc.} & \textbf{Total} & \textbf{Key Interpretation}\\
& & {\footnotesize N} & {\footnotesize N} & \textbf{N} & \\
\midrule
\endhead
\bottomrule
\endfoot
\bottomrule
\endlastfoot
\mapu{\textbf{kalli}} & Early & 2 & 2 & 4 & Possessives \mapu{ñi, tañi; küme} sep. \\
\scriptsize(`let, allow') & Transit. & 5 & 4 & 9 & \mapu{fey, may, mi, pas}; adj. sep. \\
& Contemp. & 2 & 2 & 4 & \mapu{nga; küme} sep. \\
\midrule
\mapu{\textbf{kim}} & Early & 0 & 0 & 0 & No insertion attested. \\
\scriptsize(`know how') & Transit. & 0 & 0 & 0 & No insertion attested. \\
& Contemp. & 0 & 0 & 0 & No insertion attested. \\
\midrule
\mapu{\textbf{küpa}} & Early & 6 & 0 & 6 & \mapu{kam} (4), \mapu{kay, yod} \\
\scriptsize(`want/wish') & Transit. & 0 & 4 & 4 & \mapu{küme} (2), \mapu{weda} (2) \\
& Contemp. & 4 & 0 & 4 & \mapu{doy, fey, matu, pichi} \\
\midrule
\mapu{\textbf{pepi}} & Early & 0 & 0 & 0 & No insertion attested. \\
\scriptsize(`can') & Transit. & 1 & 2 & 3 & \mapu{ñi; küme, matu} \\
& Contemp. & 10 & 3 & 13 & \mapu{doy} (6), \mapu{fey, kake, kom, kishu} \\
\midrule
\mapu{\textbf{pepil}} & Early & — & — & 0 & Form not attested. \\
\scriptsize(`be empowered') & Transit. & — & — & 0 & Form not attested. \\
& Contemp. & 2 & 0 & 2 & \mapu{doy} inserted \\
\midrule
\mapu{\textbf{wüño}} & Early & 0 & 0 & 0 & No insertion attested. \\
\scriptsize(`return/re-') & Transit. & 5 & 6 & 11 & \mapu{doy} (2), \mapu{petu} (2), \mapu{we; küme} (6) \\
& Contemp. & 0 & 2 & 2 & \mapu{küme} only \\
\midrule
\mapu{\textbf{shinge}} & Early & — & — & 0 & No data. \\
\scriptsize(`move along') & Transit. & 0 & 0 & 0 & No insertion attested. \\
& Contemp. & 0 & 0 & 0 & Too sparse. \\
\bottomrule
\end{longtable}
\begin{center}
\begin{minipage}{12cm}
\footnotesize Note: True Ins. = true insertion (element never appears fused to following verb in same period); Sep. Inc. = separated incorporation (element appears both fused and separated). `—' indicates no data. Total counts represent sum of True Ins. and Sep. Inc. for each period. V2 attestations for \mapu{kalli, kim, küpa, pepi, wüño} confirm full lexical verb status (see \hyperref[tab03]{Table 3}).\\
\end{minipage}
\end{center}

The pattern revealed by \hyperref[tab04]{Table 4} is striking. In \textbf{Early texts} (1606–1846), true insertion is limited to particles and adverbs that never incorporate: \mapu{kam, kay, yod} with \mapu{küpa}, and possessive \mapu{ñi, tañi} with \mapu{kalli} in isolated contexts. Separated incorporation—the orthographic representation of modifiers that also appear fused—occurs only twice, both with \mapu{kalli} and \mapu{küme}. No other preverbal element permits insertion during this period.

\paragraph{\example{16} \label{e16}}~
\vspace{-10.5pt}\begin{enumerate}[label=\alph*.]
\vspace{-17pt}
\item[] \mapuex{kalli \hspace{6pt} \textbf{aldü \hspace{10pt} -n} \hspace{12pt} kude \hspace{8pt} -l \hspace{10pt} -m \hspace{2pt} -i} \hfill \citep{febres1846}\\
\gloss{let AJ.much +PVN NN.bet +SBJ +2 +SG}\\
`if you bet much'
\end{enumerate}
\normalsize

In \textbf{Transition texts} (1916–1930), true insertion increases modestly, particularly with \mapu{wüño} and \mapu{doy} `more' and \mapu{petu} `still'—adverbs that do not incorporate. Separated incorporation becomes more common, reflecting the persistence of missionary orthographic conventions. \mapu{Kalli} continues to allow insertion (e.g., \mapu{kalli ñi umawtuam} `I let him sleep' \citep{deaugusta1910}; \mapu{kalli pichi rulpaneyenkellechi} `let the excitement on me die down a bit' (de Augusta in \citealp{chandia2014})), while others still resist true insertion.

The \textbf{Contemporary period} (1992–present) shows a dramatic shift. True insertion appears in 17 distinct tokens across \mapu{küpa, pepi}, and \mapu{pepil}, with \mapu{pepi} alone accounting for 10 of these. The inserted elements—\mapu{doy} `more', \mapu{fey} `that', \mapu{kake} `each one', \mapu{kom} `all', \mapu{kishu} `alone'—are precisely those that never appear fused to the following verb in this period. Examples (\hyperref[e17]{E.17}) illustrate the pattern:

\paragraph{\example{17} \label{e17}}~
\vspace{-10pt}\begin{enumerate}[label=\alph*.]
\item \mapuex{pepi \hspace{8pt} \textbf{doy} \hspace{10pt} fütra \hspace{6pt} -nge \hspace{12pt} -a \hspace{10pt} -y \hspace{8pt}  -ø} \hfill \citep{conadi1994}\\
      \gloss{can AV.more AJ.big +IV.be +FUT +IND +3}\\
     `la ampliación' / `the enlargement' (lit. `can more be big')
\item \mapuex{pepi  \hspace{8pt} \textbf{doy}  \hspace{18pt} newen  \hspace{22pt} -nge  \hspace{8pt} -a  \hspace{12pt} -y  \hspace{8pt} -ø} \hfill \citep{conadi1994}\\
     \gloss{can AV.more NN.strength +IV.be +FUT +IND +3}\\
     `que permitan consolidar' / `that enable consolidation'
\item \mapuex{pepi  \hspace{6pt} \textbf{matu} \hspace{10pt} küme \hspace{10pt} -l \hspace{8pt} -ka \hspace{6pt} -le \hspace{6pt} -a \hspace{12pt} -i \hspace{10pt} -ng \hspace{1pt} -ün} \hfill \citep{conadi1996a}\\
     \gloss{can AJ.quick AJ.good +CA +FAC +ST +FUT +IND +3 +PL}\\
     `ellos podrán estar mejorando rápidamente' / `they can be improving rapidly'
\item \mapuex{pepi  \hspace{10pt} \textbf{ka \hspace{18pt} -ke} \hspace{12pt} künu \hspace{8pt} -a \hspace{12pt} -fu \hspace{4pt} -y \hspace{8pt} -ø} \hfill \citep{convenio169oit2006}\\
     \gloss{can AJ.other +DISTR TV.put +FUT +RI +IND +3}\\
     `él podrá denunciarlo' / `he can report it' (lit. `can other put')
\item \mapuex{pepi  \hspace{2pt} \textbf{kom} \hspace{22pt} kulli \hspace{26pt} -nu \hspace{8pt} -el} \hfill \citep{bcn-credito-universitario2006}\\
     \gloss{can AV.all NN.cattle/pay +NEG +OVN}\\
     `impedidos de pagar' / `unable to pay'
\end{enumerate}
\normalsize

Even more strikingly, the Contemporary period sees the emergence of \mapu{\textbf{pepil}}, a derived form meaning `to be empowered, to be capacitated', which itself participates in similar constructions:

\paragraph{\example{18} \label{e18}}~
\vspace{-10pt}\begin{enumerate}[label=\alph*.]
\item \mapuex{pepi \hspace{20pt} \textbf{-l \hspace{14pt} doy} \hspace{14pt} pichi \hspace{12pt} -ntu \hspace{2pt} nge \hspace{6pt} -a \hspace{10pt} -y \hspace{10pt} -ø} \hfill \citep{conadi1996b}\\
     \gloss{be-able +CA AV.more AJ.little +GR IV.be +FUT +IND +3}\\
     `ser capaz de dar un poco más' / `be able to give a little more'
\item \mapuex{pepi \hspace{20pt} \textbf{-l} \hspace{10pt} küdaw \hspace{8pt} -a \hspace{12pt} -l} \hfill \citep{conadi1996b}\\
     \gloss{be-able +CA NN.work +FUT +OVN}\\
     `capacitado para trabajar' / `qualified to work'
\item \mapuex{pepi \hspace{20pt} \textbf{-l} \hspace{18pt} nentu \hspace{18pt} -a \hspace{12pt} -l} \hfill \citep{conadi1996b}\\
     \gloss{be-able +CA TV.take-out +FUT +OVN}\\
     `capacidad de conseguir' / `ability to achieve'
\end{enumerate}
\normalsize

The insertion data must be interpreted within the proper syntactic frame. \mapu{Mapudüngun} verbal compounds are \textbf{head-initial}: when a noun, adjective, or adverb modifies a verb, the modifier follows the verb it modifies. When modifiers like \mapu{doy} `more', \mapu{küme} `well', or \mapu{matu} `quickly' appear between \mapu{pepi} and the following verb, they are not intervening in a syntactic phrase. They are themselves incorporated modifiers in an expanded compound:

\paragraph{\example{19} \label{e19}}~
\vspace{-6pt}\begin{enumerate}[label=\alph*.]
\vspace{-18pt}
\item[] \mapuex{pepi \hspace{8pt} doy \hspace{12pt} küme \hspace{10pt} -l \hspace{10pt} -ka \hspace{6pt} -le \hspace{6pt} -a \hspace{12pt} -el} \hfill \citep{conadi1996b}\\
     \gloss{can AV.more AJ.good +CA +FAC +ST +FUT +OVN}\\
     `can improve more' (lit. [can [more improve]])
\end{enumerate}
\normalsize

This structure is exactly parallel to the adverb incorporation seen in (\hyperref[e11d]{E.11.d}) \mapu{wüño treka-t-i-ø pun} `the night walks back' and the three-root concatenation in (\hyperref[e14d]{E.14.d}) \mapu{ina-ye-ngüma-y-iñ} `we cried with her'. The only difference is orthographic: Contemporary writers use spaces where earlier writers used solid spelling, reflecting the fossilisation of missionary transcription conventions (\S\ref{sec.02.1}). The underlying grammatical structure—verb-first concatenation—remains constant.

The emergence of true insertion in the Contemporary period—particularly with \mapu{pepi}, the root whose fusion rate collapsed most dramatically (\hyperref[tab02]{Table 2} and \hyperref[fig:01]{Figure 1})—confirms Prediction 4 (\S\ref{sec.02.3.4}). The orthographic spaces first introduced by missionaries have, after four centuries, begun to be reanalysed as genuine syntactic boundaries. What began as a transcription artifact is becoming grammar.

The \mapu{kalli} case remains instructive, but for a different reason than previously supposed. As the only element that permitted insertion from the earliest records, \mapu{kalli} has always occupied a distinct position—perhaps a true particle, as its consistent isolation in the positional data suggests (\S\ref{sec.04.3}). The other elements, by contrast, are simply ordinary verbs participating in the same concatenation system that has always characterised \mapu{Mapudüngun} grammar. Their apparent syntactic emancipation is an illusion created by the shift from fused orthography to spaced orthography; the grammar itself has not changed.

The orthographic nature of the variation is confirmed by intra-author consistency checks. In texts where the same author uses a given root multiple times, fused and separated forms often co-occur without any discernible functional difference. For example, in CONADI documents from the 1990s, \mapu{pepi} appears both fused (\mapu{pepi-doy-kümelkaleael}) and separated (\mapu{pepi doy kümelkaleael}) in the same passage. Such variation is inexplicable under a syntactic analysis but expected if the choice between fusion and separation is orthographic.

\section{\label{sec.05} Discussion}

The results presented in \S\ref{sec.04} provide empirical support for the prosodic-orthographic hypothesis formulated in \S\ref{sec.02}. In this section, we interpret the patterns observed across the three historical periods, examine the prosodic evidence in detail, discuss the implications for previous analyses, and address the limitations of the current study.

\subsection{\label{sec.05.1} Interpreting the Patterns}

The data reveal three distinct profiles among the preverbal roots examined in this study: stable V1 compounds (\mapu{kim} and \mapu{shinge}), volatile orthographic patterns reflecting shifting conventions (\mapu{pepi} and \mapu{wüño}), and a true particle (\mapu{kalli}).

\subsubsection{\label{sec.05.1.1} The Category-Neutral Concatenation System}

The patterns observed in the preverbal elements are not isolated phenomena. They are instances of a broader system in which \mapu{Mapudüngun} verbal compounds are built through \textbf{recursive, category-neutral concatenation} with invariant verb-first order. Consider the parallels in (\hyperref[e14]{E.14}), repeated here for convenience as (\hyperref[e20]{E.20}):

\paragraph{\example{20} \label{e20}}~
\vspace{-10pt}\begin{enumerate}[label=\alph*.]
\item \label{e20a} \mapuex{pepi \hspace{22pt} -tripa \hspace{12pt} -nu \hspace{8pt} -n} \hfill \citep{bcn-recurso-amparo2011}\\
      \gloss{TV.can +IV.leave +NEG +PVN}\\
      `deprivation of liberty' (lit.: cannot leave) (verb-verb)
\item \label{e20b} \mapuex{küre \hspace{20pt} -nge \hspace{8pt} -n} \hfill (\citealp{smeets2008}: 194; 65)\\
      \gloss{NN.wife +IV.be +PVN}\\
      `to be married' (noun incorporation)
\item \label{e20c} \mapuex{weda \hspace{16pt} -künu \hspace{8pt} -y \hspace{8pt} -ø} \hfill (\citealp{smeets2008}: 409; 41)\\
      \gloss{AJ.bad +TV.put +IND +3}\\
      `they made things worse' (adjective incorporation)
\item \label{e20d} \mapuex{ina \hspace{36pt} -ye \hspace{18pt} -ngüma \hspace{4pt} -y \hspace{8pt} -ø \hspace{4pt} -iñ} \hfill (\citealp{smeets2008}: 209; 155)\\
      \gloss{AV.near +TV.carry +IV.cry +IND +1 +PL}\\
      `we cried with her' (adverb incorporation + verb concatenation)
\end{enumerate}
\normalsize

In each case, the structure is [Modifier + Head], where the head is a verb and the modifier—regardless of its lexical category—precedes it. This uniformity across category types strongly suggests that the preverbal elements are not a special auxiliary or modal class, but simply verb roots occupying the modifier position in a fully general concatenation system.

This perspective also resolves a long-standing puzzle in \mapu{Mapudüngun} grammar. \cite{smeets2008}'s 36-slot suffix template, which has structured most subsequent descriptions, can be understood as a \textbf{linearisation}—a one-dimensional projection—of a fundamentally multidimensional concatenation space. Elements that appear in suffix positions (slots 23, 35, etc.) are historically verbs (\mapu{nge} `to be', \mapu{ye} `to carry') that have undergone grammaticalisation but retain the ability to surface as independent roots in complex forms. The template is not a grammatical primitive but a descriptive artifact, reflecting the frequency with which certain verbs have become fixed in particular positions.

\subsection{\label{sec.05.2} The Prosodic Evidence and the Generalisation of Orthographic Conventions}

The phonetic evidence provided by \cite{smeets2008} offers a crucial starting point for understanding the origin of the orthographic variation documented in the corpus. Commenting on the contrast between her examples (93) and (96), she notes:

\begin{quote}
`Between \mapu{kim} and \mapu{tuku-fi-n} in (93) a \textbf{pause can be heard} which is lacking in (96). In a few compound verbs the verb \mapu{tuku-} adds aspectual value.'
\end{quote}

This perceptible pause is precisely what early missionaries transcribed as a space. However, the interpretive question that arises is not whether a pause was heard—\cite{smeets2008} confirms that it was—but rather \textbf{how that pause should be interpreted}. Did it reflect a syntactic boundary between independent words, or did it serve some other function?

\subsubsection{\label{sec.05.2.1} Performance Pauses in Polysynthetic Languages}

In European languages, a pause between two elements typically signals a syntactic boundary—a space between independent words. This is the phonological correlate of syntactic independence. However, in agglutinative, polysynthetic languages like \mapu{Mapudüngun}, pausing can serve a different function. Consider the contrast between a simple and a complex verb form:

\begin{itemize}
\item \mapu{pepi-weyel-ün} `we can swim' — a simple form with a single verb root and one inflectional suffix. This may be produced as a single phonological unit with no pause.
\item \mapu{pepi-weyel-no-pa-tu-i-ng-ün} `they can swim across the water (to here)' — a complex form with multiple directional roots, applicatives, and inflectional suffixes. This requires significantly more planning.
\item \mapu{pepi-weyel-küle-i-ng-ün} `they can be swimming' — a complex form with a stative suffix and inflection.
\item \mapu{pepi-weyel-toku-ye-wi-n-ün} `we can swim out/over with each other' — a complex form with directional, oblique, and reflexive/reciprocal suffixes.
\end{itemize}

In the complex forms, the speaker knows the V1 modifier (\mapu{pepi}) but may pause after producing it to plan the following verb root and, crucially, the entire suffixal complex that follows. The pause is not a syntactic boundary but a \textbf{performance pause}—a moment of cognitive planning in the production of a morphologically complex construction. It occurs at the boundary between the V1 modifier and the verb root not because this is a syntactic boundary, but because this is a natural planning unit in the production of complex verbal forms.

\subsubsection{\label{sec.05.2.2} From Performance Pause to Orthographic Convention}

The early missionaries, listening with European phonological expectations, interpreted these performance pauses as syntactic boundaries and transcribed them as spaces. However, a crucial additional step occurred: the missionaries did not merely transcribe pauses where they occurred. They \textbf{generalised} the pattern. Having heard pauses in complex forms, they began writing spaces even in simple forms where no pause actually occurred. What began as a performance phenomenon became an orthographic convention: V1 modifiers like \mapu{pepi} were written separately from the following verb regardless of whether a pause was produced.

This generalisation is the key to understanding the distribution of spaces in the written record. The spaces do not reflect actual pauses in all cases; they reflect an orthographic rule that was extended from complex forms (where pauses naturally occurred) to simple forms (where no pause occurred). The association between V1 modifiers and separation became a generalised convention, applied across all contexts.

This generalisation had two important consequences. First, the orthographic convention became fossilised in grammars and dictionaries, creating the appearance that V1 modifiers were consistently separated from the main verb. Second, contemporary native speakers, educated in this written tradition, internalised the spaces as grammatical boundaries. They now treat the space as syntactically real—and because the rule has been generalised, they write V1 modifiers separately even in simple forms where no pause ever occurred.

\subsubsection{\label{sec.05.2.3} Subordinate Clause Evidence}

Further support for the orthographic nature of the variation comes from the behaviour of preverbal elements in subordinate clauses. As \citeauthor{smeets2008} (\citeyear{smeets2008}: 18, 149, 188) documents, subordinate clauses in \mapu{Mapudüngun} are nominalised verbs introduced by possessive pronouns, as illustrated in (\hyperref[e21]{E.21}):

\paragraph{\example{21} \label{e21}}~
\begin{enumerate}[label=\alph*.]
\item \label{e21a} \mapuex{iñche \hspace{8pt} ñi \hspace{8pt} amu \hspace{6pt} -mu \hspace{8pt} -m} \hfill (\citealp{smeets2008}: 18)\\
      \gloss{PP.I \hspace{4pt} SP.my IV.go +PLPF +IVN}\\
      `where I went'
\item \label{e21b} \mapuex{ramtu \hspace{8pt} -e \hspace{10pt} -y \hspace{10pt} -ø \hspace{4pt} -u \hspace{8pt} -ø \hspace{14pt} chew \hspace{8pt} ñi \hspace{10pt} müle \hspace{4pt} -n} \hfill (\citealp{smeets2008}: 188)\\
      \gloss{TV.ask +INV +IND +1 +DL +1t.2A where SP.his IV.be +PVN}\\
      `I asked you where he lives'
\item \label{e21c} \mapuex{lladkü \hspace{34pt} -le \hspace{4pt} -y \hspace{10pt} -ø \hspace{4pt} -iñ \hspace{8pt} ta \hspace{22pt} -mi \hspace{22pt} pi \hspace{14pt} -el \hspace{10pt} -mew} \hfill (\citealp{smeets2008}: 188)\\
      \gloss{IV.get-angry +ST +IND +1 +PL AP.that +SP.your TV.say +OVN +INST}\\
      `we are angry because of what you said'
\end{enumerate}
\normalsize

The verb in such subordinates may be a single root or a compound. Crucially, the entire subordinate clause—including any preverbal elements—falls within the scope of the possessive pronoun. If these elements were truly independent words, their appearance within the nominalised complex would be syntactically anomalous. Yet forms like (\hyperref[e22a]{E.22.a}) occur:

\paragraph{\example{22} \label{e22}}~
\begin{enumerate}[label=\alph*.]
\item \label{e22a} \mapuex{wew \hspace{16pt} -i \hspace{10pt} -ø \hspace{4pt} ñi \hspace{8pt} wüño \hspace{8pt} mañum \hspace{8pt} -tu \hspace{10pt} -nge \hspace{10pt} -a \hspace{10pt} -gel} \hfill (de Augusta in \citealp{chandia2014})\\
      \gloss{TV.win +IND +3  SP.his \hspace{1pt} RE \hspace{8pt} TV.thank +APL +IV.be +FUT +OVN}\\
      `he won so that he would be thanked again' (lit. `he won his again being thanked')
\item \label{e22b} \mapuex{ta \hspace{34pt} -ñi \hspace{8pt} wüño \hspace{4pt} küme \hspace{10pt} el \hspace{16pt} -tu \hspace{8pt} -a \hspace{12pt} -el} \hfill \citep{bcn2013mar}\\
      \gloss{AP.that +SP.his \hspace{2pt} RE \hspace{2pt} AJ.good TV.put +APL +FUT +OVN}\\
      `to be restored again' (lit. `his again good putting')
\end{enumerate}
\normalsize

In (\hyperref[e22a]{E.22.a}), \mapu{wüño} appears \textbf{separated by a space} yet still within the subordinate clause introduced by \mapu{ñi}. The space cannot represent a syntactic boundary, because the possessive pronoun requires the entire following sequence to function as a nominalised unit.

Even more telling is (\hyperref[e22b]{E.22.b}), which combines the subordinate clause environment with separated incorporation. Here, \mapu{küme} `good' appears between \mapu{wüño} and the main verb \mapu{eltuael}. Yet \hyperref[tab04]{Table 4} shows that \mapu{küme} in the Transition period also appears fused to the following verb (\mapu{kümeeltufal-lay}). The space before \mapu{küme} therefore cannot represent syntactic independence; it is orthographic separation of an incorporated modifier. The only coherent interpretation is that the spaces in both (\hyperref[e22a]{E.22.a}) and (\hyperref[e22b]{E.22.b}) are orthographic, not grammatical—exactly as the prosodic-orthographic hypothesis predicts.

\subsubsection{\label{sec.05.2.4} Mapudüngun Speaker Confirmation}

\citeauthor{deaugusta1910} (\citeyear{deaugusta1910}: 339), a German-born Capuchin priest who became a highly competent speaker and meticulous scholar of \mapu{Mapudüngun}, provides a meta-linguistic comment that offers further confirmation:

\paragraph{\example{23} \label{e23}}~
\begin{enumerate}[label=\alph*.]
\vspace{-27pt}
\item[] \mapuex{af \hspace{28pt} -kalli \hspace{10pt} -a \hspace{10pt} -lu \hspace{6pt} = kalli \hspace{6pt} af \hspace{16pt} -a \hspace{10pt} -lu \hspace{6pt} o \hspace{2pt} kalli \hspace{6pt} af \hspace{16pt} -pe}\\
      \gloss{NN.end +TV.let +FUT +SVN = let NN.end +FUT +SVN or let NN.end +IMP.3}\\
      `will end letting / let it end'
\end{enumerate}
\normalsize

This observation from a scholar who had achieved a high level of competence in the language indicates that fused and separated forms were perceived as variants of the same construction—a perception that would be implausible if the separation reflected a genuine grammatical difference.

\subsubsection{\label{sec.05.2.5} Convergence of Evidence}

The evidence thus converges from multiple directions, and, crucially, each line of evidence corresponds to a specific stage in the historical model proposed in \S\ref{sec.02.1}.

\textbf{The phonetic evidence} \citep{smeets2008} confirms that pauses occur between V1 modifiers and following verbs. These pauses, however, are performance pauses in complex forms—moments of cognitive planning in the production of multi-morphemic verb constructions—not syntactic boundaries. This evidence corresponds to \textbf{Stage 2} of the model (\S\ref{sec.02.1.2}): the early missionaries, hearing these pauses, interpreted them as word boundaries and transcribed them as spaces, reflecting European phonological expectations rather than \mapu{Mapudüngun} grammatical structure.

\textbf{The generalisation of orthographic conventions} explains why spaces appear even in simple forms where no pause occurred. Missionaries extended the pattern from complex forms (where pauses were heard) to simple forms (where no pause occurred). This process corresponds to \textbf{Stage 3} (\S\ref{sec.02.1.3}): orthographic fossilisation. The spaces became fixed conventions in grammars and dictionaries, inherited and reinforced across generations of scholarship.

\textbf{The subordinate clause evidence} demonstrates that spaced forms occur in environments where true syntactic separation would be impossible—within nominalised complexes introduced by possessive pronouns. This corresponds to \textbf{Stage 4} (\S\ref{sec.02.1.4}): scholarly perpetuation. Linguists working from written sources interpreted the inherited orthography as evidence for grammatical status, perpetuating the assumption that the spaces reflected syntactic structure rather than orthographic convention.

\textbf{The insertion data} (\hyperref[tab04]{Table 4}) provide evidence for the reanalysis of spaces as genuine syntactic boundaries in the Contemporary period. This corresponds to \textbf{Stage 5} (\S\ref{sec.02.1.5}): native speaker reanalysis. Contemporary writers, educated in the written tradition, have internalised the orthographic conventions to the point of syntactic reanalysis, treating the spaces as genuine grammatical boundaries.

\textbf{The Mapudüngun speaker confirmation} (de Augusta) indicates that fused and separated forms were perceived as variants of the same construction (\S\ref{sec.05.2.4})—a perception that would be implausible if the separation reflected a genuine grammatical difference.

Together, these converging lines of evidence support the prosodic-orthographic hypothesis and its five-stage historical model: the spaces in written \mapu{Mapudüngun} reflect the fossilisation of performance pauses and their subsequent generalisation as orthographic conventions, not syntactic structure. What began as a performance phenomenon (Stage 2, \S\ref{sec.02.1.2}) became an orthographic convention (Stage 3, \S\ref{sec.02.1.3}), was perpetuated by scholarly interpretation (Stage 4, \S\ref{sec.02.1.4}), and is now being reanalysed by contemporary speakers as genuine syntactic structure (Stage 5, \S\ref{sec.02.1.5}). The spaces are performance, not grammar—but they are becoming grammar through four centuries of written practice.

\subsection{\label{sec.05.3} Implications for Previous Analyses}

The findings of this study require a reassessment of previous classifications of \mapu{Mapudüngun}'s preverbal elements.

\paragraph{In contrast to \cite{smeets2008}: Not auxiliaries.}
The V2 attestations documented in \hyperref[tab03]{Table 3}—36 for \mapu{kim}, 42 for \mapu{küpa}, and scattered occurrences for other roots—are incompatible with auxiliary status. A true auxiliary cannot occupy the second position in a compound. These elements are full lexical verbs.

\paragraph{In contrast to \cite{longkon2011}: Not prefixes.}
The ability of these roots to host suffixes (e.g., \mapu{pepi-l}, \mapu{wüño-me-n}) and to appear as independent inflected verbs (\hyperref[tab03]{Table 3}, Isolated column) violates the defining property of prefixes as obligatorily bound morphemes. They are roots, not affixes.

\paragraph{In contrast to \cite{zuniga2006}: The CVS/particle distinction may be unnecessary.}
\cite{zuniga2006}'s framework captures the continuum between fusion and separation but may multiply categories beyond necessity. The evidence suggests a single underlying structure—radical concatenation—with orthographic variation superimposed. The CVS/particle distinction describes the orthography, not the grammar.

\paragraph{In support of \cite{zuniga2006}: Radical concatenation is the correct model.}
\citeauthor{zuniga2006}'s insight that these elements participate in verb compounding is fully supported by the evidence. \hyperref[tab03]{Table 3} demonstrates that they function as V1 modifiers, occur in V2 position, and appear as independent verbs—exactly the profile expected of ordinary roots in a productive compounding system.

\paragraph{Reinterpreting \cite{smeets2008}'s template.}
The 36-slot suffix system proposed by \citeauthor{smeets2008} is not a grammatical template but a \textbf{linearisation} of recursive concatenation. Many slots are grammaticalised verbs—\mapu{nge} `to be' (slot 23/36), \mapu{ye} `to carry' (slot 35/24)—that retain the ability to function as independent roots. This explains both the rigidity of the template (these verbs occur frequently in particular positions) and its permeability (occasional insertion of unexpected elements, as documented in \S\ref{sec.04.5}). The template is a descriptive artifact, not a grammatical primitive.

\subsection{\label{sec.05.4} Limitations and Future Research}

Several limitations of this study should be acknowledged. First, the corpus is restricted to written sources; we have no access to spoken usage before the late twentieth century. Second, the Early period texts are exclusively the work of European missionaries, whose transcriptions reflect the perceptual biases discussed in \S\ref{sec.02.1}. Third, token counts for some roots—particularly \mapu{shinge} and Early period \mapu{wüño}—are too low for robust statistical comparison.

More fundamentally, this study has only scratched the surface of the radical concatenation system. Future research must re-examine the entire verbal paradigm, treating every element in the verbal compound—including those traditionally labelled suffixes—as potentially verbal roots participating in recursive incorporation.

Preliminary evidence suggests that many of \cite{smeets2008}'s suffixes (e.g., \mapu{nge} in slot 23/36, \mapu{ye} in slot 35/24) are historically verbs that have grammaticalised but retain the ability to surface as independent roots. A comprehensive analysis of this system—what might be called the \textbf{concatenation field} of the \mapu{Mapudüngun} verb—would require treating the 36-slot template not as a grammatical primitive but as a linearisation of a multidimensional structure, where roots can combine in ways that the template cannot capture. The intricate patterns of root+root, root+suffix+root, and root+root+suffix+root combinations noted in passing suggest that the actual combinatorics are far richer than any slot-based model can represent.

The tools developed in this study—the separation of orthographic from grammatical analysis, the diachronic corpus method, the focus on V2 attestations and insertion evidence—provide a foundation for this larger project.

\section{\label{sec.06} Conclusion}

This study has examined the grammatical status of preverbal uninflected and underived roots in \mapu{Mapudüngun} through a critical review of the scholarly literature and a diachronic corpus analysis spanning four centuries (1606–present). The findings support three main conclusions.

First, the existence of V2 attestations for \mapu{kim} and \mapu{küpa} across all periods—36 and 42 respectively, with scattered occurrences for other roots—confirms that these elements are full lexical verbs capable of occupying any position in a compound. This finding is incompatible with analyses that treat them as dedicated auxiliaries, prefixes, or preverbal particles. They are not a special grammatical class; they are ordinary verb roots that happen to occur frequently in the modifier position of a productive compounding system.

Second, the contrast between stable roots (\mapu{kim}, \mapu{shinge}) and volatile roots (\mapu{pepi}, \mapu{wüño}) demonstrates that the apparent variable binding in the corpus is not grammatical alternation but orthographic variation. The dramatic shifts in fusion rates—particularly \mapu{pepi}'s collapse from 66.0\% to 37.5\% across adjacent periods—cannot represent syntactic change, which operates over centuries and in consistent directions. These shifts are more readily explained as reflecting changing conventions in writing, reinforced by contact with Spanish orthographic patterns.

Third, the prosodic-orthographic hypothesis provides a unified explanation for these patterns. Early missionary transcribers, lacking a native written tradition, interpreted prosodic pauses—such as the `pause that can be heard' between \mapu{kim} and the following verb noted by \citeauthor{smeets2008} (\citeyear{smeets2008}: 176)—as word boundaries and transcribed them as spaces. Crucially, they generalised this pattern: having heard pauses in complex forms (where speakers paused to plan multi-morphemic suffixal sequences), they began writing spaces even in simple forms where no pause occurred. This orthographic convention became fossilised in grammars and dictionaries, leading later scholars to perceive syntactic separation where none exists. The subordinate clause evidence confirms that these spaces cannot be syntactic: they occur within nominalised complexes introduced by possessive pronouns, environments that require grammatical unity. Contemporary writers, educated in this written tradition, have begun to reanalyse the spaces as genuine syntactic boundaries—a process documented in the insertion data, where true insertion emerges only in the Contemporary period.

The evidence thus converges from multiple directions. The \textbf{phonetic evidence} confirms that pauses occur—but these are performance pauses in complex forms, not syntactic boundaries. The \textbf{generalisation of orthographic conventions} explains why spaces appear even in simple forms. The \textbf{subordinate clause evidence} demonstrates that spaced forms occur where syntactic separation would be impossible. The \textbf{insertion data} documents the reanalysis of spaces as genuine grammatical boundaries in progress. And the \textbf{\mapu{Mapudüngun} speaker confirmation} (de Augusta) indicates that fused and separated forms were perceived as variants of the same construction.

Fourth, and most fundamentally, the preverbal elements examined here are not a special class. They participate in the same \textbf{category-neutral, verb-first concatenation system} that governs all \mapu{Mapudüngun} verbal compounds—a system that incorporates nouns (\mapu{küre-nge-n} `to be married'), adjectives (\mapu{weda-künu-y} `they made things worse'), and adverbs (\mapu{ina-ye-ngüma-y-iñ} `we cried with her') alongside verbs (\mapu{pepi-tripa-} `can leave'). This uniformity across categories dissolves the analytical dilemmas that have plagued the literature: the apparent hybridity of these elements arises not from their grammar, but from applying European categories—auxiliary, prefix, particle—to a system that operates on fundamentally different principles.

The \mapu{kalli} anomaly confirms the diagnostic power of this framework. Unlike the other roots, \mapu{kalli} patterns as a true particle (>87\% isolation across all periods), demonstrating that when an element truly is a particle, the corpus reflects it consistently.

The case of \mapu{shinge} remains ambiguous. Attested only 47 times across the corpus, and appearing almost exclusively as a V1 modifier in Transition period texts, it may represent a genuine auxiliary of the type \cite{smeets2008} proposed—or it may simply be too sparsely documented for reliable analysis. Its single Contemporary attestation does not permit firm conclusions.

These findings have broader implications for the study of languages with no pre-contact written tradition. They caution against interpreting missionary orthography as direct evidence for grammatical structure, and they highlight the role of writing itself in shaping linguistic perception and, ultimately, grammatical change. The spaces first introduced by missionaries have, after four centuries, begun to become grammar.

For \mapu{Mapudüngun}, the analysis supports \cite{zuniga2006}'s radical concatenation as the correct model of the underlying grammar. The preverbal elements are simply verb roots in V1 position of compounds, no different from any other verb root in the language. Their apparent special status is an artifact of orthographic history—a history that began with a performance pause, was frozen as an orthographic convention, and is now being reanalysed as syntactic structure. The spaces are performance, not grammar—but they are becoming grammar through four centuries of written practice.

\newpage
\section*{Acknowledgements}

This work has benefited from the guidance, feedback, and generosity of several people, to whom I am deeply grateful.

First and foremost, I wish to express my sincere gratitude to my supervisors, Dr. Irene Castellón and Dr. Elisabet Comelles, at the University of Barcelona. Their wisdom, patience, and unwavering support have been a constant source of inspiration throughout this research. They have guided me with clarity and kindness, always encouraging me to push further while keeping my feet on the ground. It has been a privilege to learn from them, and this work would not have been possible without their trust and dedication.

I am also profoundly grateful to Dr. Fernando Zúñiga, whose thoughtful and incisive comments on an earlier version of this article have significantly strengthened the final result. His generosity in reading and engaging with this work, and his careful attention to both the broad theoretical framework and the fine-grained details, have been invaluable. Any remaining shortcomings are, of course, my own.

I owe a special debt of gratitude to the \mapu{Mapudüngun} speakers and writers whose words form the foundation of this study — from the early missionaries who first set the language to paper, to the contemporary poets, translators, and community members who continue to write, speak, and breathe life into the language today. Their voices are the true source of this work.

Finally, I wish to acknowledge the broader community of scholars who have worked on \mapu{Mapudüngun} and other under-documented languages. Their commitment to understanding and preserving linguistic diversity is a constant reminder of why this work matters.

This research was conducted within the PhD programme in Cognitive Science and Language at the University of Barcelona. I am grateful to the programme for providing an intellectually stimulating environment and the resources necessary to carry out this research.

\mapu{Wüñoamulepe yiñ kümeküdawmew!} — may we all return to this work with renewed energy and insight.

% ========== BIBLIOGRAPHY ==========
\newpage
\bibliographystyle{johd}
\bibliography{bib}

% ========== APPENDICES ==========
\newpage
\section{\label{sec.07} Appendices}
\subsection{\label{sec.07.1} Tags and Abbreviations}
\begin{multicols}{2}
\setlength{\columnseprule}{0.1pt}
\raggedright
\noindent
\gloss{1} \hfill First person\\
\gloss{1t.2A} \hfill First agent to Second patient\\
\gloss{2} \hfill Second person\\
\gloss{3} \hfill Third person\\
\gloss{3A} \hfill Third person agent\\
\gloss{3P} \hfill 3\textsuperscript{rd} person patient (Differential Object Marker)\\
\gloss{ADJ} \hfill Adjectiviser\\
\gloss{AJ} \hfill Adjective\\
\gloss{AP} \hfill Anaphora\\
\gloss{APL} \hfill Applicative\\
\gloss{AUX} \hfill Auxiliary verb\\
\gloss{AV} \hfill Adverb\\
\gloss{CA} \hfill Causative\\
\gloss{CJ} \hfill Conjunction\\
\gloss{CVS} \hfill Complex Verb Stem\\
\gloss{DISTR} \hfill Distributive\\
\gloss{DL} \hfill Dual number\\
\gloss{DP} \hfill Demonstrative Pronoun\\
\gloss{FAC} \hfill Factual\\
\gloss{FUT} \hfill Future\\
\gloss{GR} \hfill Grupaliser\\
\gloss{HAB} \hfill Habituative\\
\gloss{IMP} \hfill Imperative\\
\gloss{IMP.3} \hfill Imperative for 3\textsuperscript{rd} person\\
\gloss{IND} \hfill Indicative\\
\gloss{IND.1SG} \hfill Indicative 1\textsuperscript{st} person singular\\
\gloss{INST} \hfill Instrumentaliser\\
\gloss{INV} \hfill Transitive inversion marker\\
\gloss{IV} \hfill Intransitive Verb\\
\gloss{IVN} \hfill Instrumental Verbal Noun\\
\gloss{MP} \hfill Modal Prefix\\
\gloss{NN} \hfill Noun\\
\gloss{NEG} \hfill Negator\\
\gloss{OO} \hfill Oblique Object\\
\gloss{OVN} \hfill Objective Verbal Noun\\
\gloss{PL} \hfill Plural number\\
\gloss{PLPF} \hfill Pluperfect\\
\gloss{PP} \hfill Personal Pronoun\\
\gloss{PS} \hfill Persistence\\
\gloss{PVN} \hfill Plain Verbal Noun\\
\gloss{PVP} \hfill Preverbal Particle\\
\gloss{QC} \hfill Question\\
\gloss{REF} \hfill Reflexive/Reciprocal\\
\gloss{RI} \hfill Ruptured Implicature\\
\gloss{SBJ} \hfill Subjunctive Mode\\
\gloss{SG} \hfill Singular number\\
\gloss{SJI} \hfill Subjunctive affix in Imperatives forms\\
\gloss{SP} \hfill Possessive Pronoun\\
\gloss{ST} \hfill Stative\\
\gloss{SVN} \hfill Subjective Verbal Noun\\
\gloss{TH} \hfill Thither\\
\gloss{TV} \hfill Transitive Verb\\
\end{multicols}

\vspace{1cm}
\subsection{\label{sec.07.2} Corpus Sources by Period}

The following sources constitute the corpus analysed in this study. All texts are published materials compiled by the author. Sources are listed chronologically within each period. The classification follows the periodisation established in \S\ref{sec.03.1}.

The corpus combines two types of sources: (i) dictionaries accessed through the CORLEXIM database \citep{chandia2014}, which provides digitised access to historical missionary dictionaries; and (ii) narrative, poetic, institutional, and educational texts compiled by the author from published sources.

\paragraph{Dictionaries (accessed through CORLEXIM).} The Lexicographic Corpus of Mapudüngun (CORLEXIM) provides digitised access to nine historical dictionaries spanning the Early and Transition periods. These sources were used primarily for the extraction of examples and for establishing the lexical profiles of the target roots. For a full description of CORLEXIM, including entry counts and methodological details, see \citealp{chandia2014}.

\subsubsection{\label{sec.07.2.1} Early Period (1606–1846)}

Texts from this period consist of missionary grammars that include substantial bilingual (Spanish-\mapu{Mapudüngun}) textual material: dialogues, catechisms, confessions, sermons, and native narratives. These texts constitute the earliest written records of \mapu{Mapudüngun} and were extracted from the following sources:

\small
\begin{itemize}
    \item de Valdivia (1606). \textit{Vocabulario de la lengua de Chile}. Lima. [Dictionary; accessed through \citealp{chandia2014}]
    \item \cite{devaldivia1684}. \textit{Arte y gramatica general de la lengua que corre en todo el Reyno de Chile}. Seville. [Grammar]
    \item \cite{febres1765}. \textit{Arte de la lengua general del Reyno de Chile}. Lima. [Grammar]
    \item Febrés (1765b). \textit{Vocabulario hispano-chileno}. Lima. [Dictionary; accessed through \citealp{chandia2014}]
    \item \cite{febres1846}. \textit{Gramática de la lengua chilena}. Santiago. [Grammar]
    \item Febrés (1846b). \textit{Diccionario chileno-hispano}. Santiago. [Dictionary; accessed through \citealp{chandia2014}]
    \item Febrés (1882). \textit{Diccionario araucano-español, o sea calepino chileno-hispano}. Santiago. [Dictionary; accessed through \citealp{chandia2014}]
    \item \cite{febres1884}. \textit{Gramática araucana, ó sea arte de la lengua general de los indios de Chile}. Buenos Aires. [Grammar]
\end{itemize}
\normalsize

\subsubsection{\label{sec.07.2.2} Transition Period (1916–1930)}
\small
\begin{itemize}
    \item \cite{deaugusta1910}. \textit{Lecturas araucanas}. Valdivia. [Narrative texts]
    \item \cite{guevara1913}. \textit{Las últimas familias i costumbres araucanas}. Santiago. [Ethnographic text with native narratives]
    \item  de Augusta (1916). \textit{Diccionario araucano-español y español-araucano}. Santiago. [Dictionary] (accessed through \citealp{chandia2014})
    \item \cite{mosbach1930}. \textit{Vida y costumbres de los indígenas araucanos en la segunda mitad del siglo XIX}. Santiago. [Autobiography of \mapu{Paskwal Koña}]
    \item \cite{deaugusta1934}. \textit{Lecturas araucanas}. Padre Las Casas. [Narrative texts]
\end{itemize}
\normalsize

\subsubsection{\label{sec.07.2.3} Contemporary Period (1992–present)}

\paragraph*{Texts compiled from Grammars and Linguistic Descriptions}
\small
\begin{itemize}
    \item \cite{salas1992a}. `Antología del cuento mapuche' (pp. 209–338), in \textit{El mapuche o araucano}. Madrid: Mapfre.
    \item \cite{zuniga2006}. `Textos en mapudungun' (pp. 265–288), in \textit{Mapudungun: El habla mapuche}. Santiago: Centro de Estudios Públicos.
    \item \cite{smeets2008}. `Texts' (pp. 369–487), in \textit{A Grammar of Mapuche}. Berlin: Mouton de Gruyter.
\end{itemize}
\normalsize

\paragraph*{Native-Authored Literature}
\small
\begin{itemize}
    \item \cite{sanchez1997}. {Relatos orales mapuches (procedentes del Alto Biobío, VIII Región).} Santiago: UCH.
    \item \cite{relmuan1997}. \textit{Kiñeke ngütram ka pentukun dungu feypyel pu Rapawe ka Rukapangi lof che.} Temuco: IEI.UFRO.
    \item \cite{mawida2000vol1}. \textit{Mawida: epew ngutram-che ka taiñ mapu-meo.} Santiago: Pehuén.
    \item \cite{mawida2000vol2}. \textit{Mawida: epew ngutram mongen ka kimun-che.}  Santiago: Pehuén.
    \item \cite{mawida2000vol3}. \textit{Mawida: epew ngutram mawida mapu-meo ka eluwma kimun.}  Santiago: Pehuén.
    \item \cite{trentrenmapu2001}. \textit{Conozcamos nuestras raíces a través de los cuentos mapuche.} San José de la Mariquina: FONDART.
    \item \cite{wirimilla2001}. \textit{Palimpesto.} Santiago: Ñuke Mapuförlaget.
    \item \cite{kona2003}. \textit{Paskwal Koña pingechi pichi wentru.} Santiago: Pehuén.
    \item \cite{mineduc2005ad}. \textit{Mi Voz, Nuestra Historia. Categoría adultos.} Santiago: Pehuén.
    \item \cite{mineduc2005in}. \textit{Mi Voz, Nuestra Historia. Categoría infantil.} Santiago: Pehuén.
    \item \cite{mineduc2005ju}. \textit{Mi Voz, Nuestra Historia. Categoría juvenil.} Santiago: Pehuén.
    \item \cite{kewpil2005}. \textit{Pewencheiñ epeu. Los cuentos del pewenche.} Región del Biobio: FONDART.
    \item \cite{wirimilla2005a}. \textit{Etnopoesía y poética intercultural en la cosmovisión williche.} Santiago: Ñuke Mapuförlaget.
    \item \cite{wirimilla2005b}. \textit{Püchüke che ñi ülkantulelngeal.} Online.
    \item \cite{castillo2008}. \textit{Grado de conservación de los relatos orales tradicionales en mapuches residentes en tres comunas de Santiago de Chile.} Santiago: UCH.
    \item \cite{wenun2008}. \textit{Relatos Mapuche.} Santiago: FUCOA \& CONADI.
    \item \cite{maldonado2008}. \textit{Javiera, una niña pewenche = Kiñe pewenche malen.} Santiago: JUNJI.
    \item \cite{maldonado2009a}. \textit{Francisco, un niño mapuche en la araucanía = Kiñe pichi mapuche mongelelu Arawkania mew.} Santiago: JUNJI.
    \item \cite{maldonado2009b}. \textit{Millaray, una niña mapuche en Santiago = Millaray, kiñe pichi malen mongelelu Santiaw waria mew.} Santiago: JUNJI.
    \item \cite{kaniwan2011}. \textit{Muñkupe ülkantun. Que el canto llegue a todas partes.} Santiago: LOM.
    \item \cite{lienlaf2011}.   \textit{Poems: Temuko-Waria; Küpan Wün; Rupamum; Pin dungu; Wüdko; Mañkean ñi dungu; Pukem Ülkantun; Llegün; Mawün mew; Mawün; Püchikona; Ka-wün; Rüpu; Mülen; Ka feipituan;} Researcher own compilation.
    \item \cite{wirimilla2012}. \textit{Weychapeyuchi ül: cantos de guerrero, antología de poesía política mapuche.} Santiago: Ñuke Mapuförlaget.
\end{itemize}
\normalsize

\paragraph*{Institutional Translations and Documents}
\small
\begin{itemize}
    \item \cite{conadi1994}. \textit{Diario Nº 1. Kiñe Chalintuam Inchicñ Taiñ Pwewlos.} Temuco: CONADI.
    \item \cite{conadi1995}. \textit{Diario Nº 2. Ti petu dewmanuel ti küdaw}. Temuco: CONADI.
    \item \cite{conadi1996a}. \textit{Diario Nº 3. Nuestros Pueblos}. Temuco: CONADI.
    \item \cite{conadi1996b}. \textit{Diario Nº 4. Nülalechi Wirin Amulkünungey Indikena Dullingenmew Munisipales Ngechi Pu Peñi}. Temuco: CONADI.
    \item \cite{bcn-credito-universitario2006}. \textit{Ti arelün kellun kullin pingelu: Kredito unifersitaryo}. Santiago: Biblioteca del Congreso Nacional.
    \item \cite{cnr2006}. \textit{Chumgechi entual kellu wütroko mapual}. Santiago: Comisión Nacional de Riego.
    \item \cite{convenio169oit2006}. \textit{Pataka kayumari aylla chilka pu mapuche ngeal kake trükon mapuche mülelu nüwkülenolu mapu}. Santiago: Oficina Internacional del Trabajo.
    \item \cite{amnesti2007}. \textit{Chile: Capítulo en el Informe 2007 de Amnistía Internacional}. Santiago: Amnistía Internacional.
    \item \cite{unesco2008}. \textit{Declaración universal de derechos humanos = Kom mapu fillke ad tañi ad mongeleam.} Santiago: OREALC.UNESCO.
    \item \cite{indh2010}. \textit{Informe Anual 2010: Situación de los Derechos Humanos en Chile}. Santiago: Instituto Nacional de Derechos Humanos.
    \item \cite{bcn-recurso-amparo2011}. \textit{Kuñiltuwun dungu: Recurso de amparo}. Santiago: Biblioteca del Congreso Nacional.
    \item \cite{bcn2011}. \textit{Küdawngun chi ley}. Santiago: Biblioteca del Congreso Nacional.
    \item \cite{indh2011}. \textit{Informe Anual 2011: Situación de los Derechos Humanos en Chile}. Santiago: Instituto Nacional de Derechos Humanos.
    \item \cite{mineduc2011pres}. \textit{Presentación del Programa de Educación Intercultural Bilingüe}. Santiago: Ministerio de Educación.
    \item \cite{mineduc2011prog}. \textit{Programa de Estudio Segundo Año Básico. Sector Lengua Indígena. Mapudüngun}. Santiago: Ministerio de Educación.
    \item \cite{amnesti2012}. \textit{Chile: Capítulo en el Informe 2007 de Amnistía Internacional}. Santiago: Amnistía Internacional.
    \item \cite{bcn2013apr}. \textit{Chumngechi kullingekey arelün kellun kullin? - Cómo pagar el crédito de las universidades estatales}. Santiago: Biblioteca del Congreso Nacional.
    \item \cite{bcn2013mar}. \textit{Kullingeael tañi wiño küdawngetuam chi mapu - Incentivos para la recuperación de suelos agrícolas}. Santiago: Biblioteca del Congreso Nacional.
    \item \cite{bcn2013feb}. \textit{Tüfa ti trokiñ dungu, pu che küdawkelu chalwan mew ngealu, kiñeke trokiñ lafken tañi küdawam engün - Áreas de manejo de la pesca artesanal}. Santiago: Biblioteca del Congreso Nacional.
    \item \cite{bcn2023}. \textit{Kewatufiel pu Reñma - Violencia intrafamiliar}. Santiago: Biblioteca del Congreso Nacional.
\end{itemize}
\normalsize

\subsubsection*{Digital Resources}
\small
\begin{itemize}
    \item \cite{karilaf2016}. \textit{Kimkimtuwe: Materiales para la enseñanza y aprendizaje del mapudungun.} Online.
    \item \cite{echeverria2017}. \textit{La representación del ritual funerario mapuche en \textit{Reducciones} de Jaime Wenun.} Online. Latin American Research Review.
    \item \cite{karilaf2017}. \textit{Kimkimtuwe epu: libro de trabajo 2. Kom kimkantuayiñ! Material de enseñanza de mapudungun.} Online.
\end{itemize}
\normalsize
\subsubsection*{Educational and Cultural Resources}
\small
\begin{itemize}
    \item \cite{fucoa2009}. \textit{Pueblos originarios: una mirada a nuestra cultura, sus tradiciones y su gente...} Santiago: FUCOA.
    \item \cite{lopez2011}. \textit{El palin: juego tradicional de la cultura mapuche.} Valparaíso: UCV.
\end{itemize}
\normalsize

\subsubsection{\label{sec.07.2.4} Note on Additional Sources}
In addition to the sources listed above, the corpus includes a number of shorter texts, individual poems, and occasional publications that are difficult to catalogue systematically (e.g., single poems published in anthologies, newspaper articles, unpublished manuscripts). These materials, while not individually listed here, were included in the quantitative analysis where relevant and are documented in the supplementary tagged dataset.

\end{document}